# Design, modelling and experimental validation of single and multisloped bipenniform architecture-based shape memory alloy-driven linear actuator

Kanhaiya Lal Chaurasiya[a] , Ruchira Kumar Pradhan[a] , Yashaswi Sinha[a] , Shivam Gupta[a] , Ujjain Kumar Bidila[b] , Digambar Killedar[b] , Kapil Das Sahu[b] and Bishakh Bhattacharya [a,*]

[a]*Indian Institute of Technology Kanpur, Kanpur, Uttar Pradesh, 208016, India*
[b]*Johnson Controls India Private Limited, Pune, Maharashtra, 411006, India*

## Abstract

Integrating bio-inspired design principles into new product development could bring forth the next generation of adaptive and energy efficient smart material-based actuation systems. The increasing industrial demand for alternative actuators over conventional electromagnetism-based systems having limited efficiency, bulky size, complex design due to in-built gear-train mechanisms, and high production and amortization costs necessitates the innovation in new actuator development. The present study explores and integrates the advantages of bipenniform architecture, which generates high force in the given physiological region and a high power-to-weight ratio of shape memory alloy (SMA), into a novel bio-inspired SMA-based linear actuator. The primary configuration investigated in this work employs a single sloped bipenniform arrangement of the SMA wires. A mathematical model of a multi-layered bipenniform configuration-based SMA actuator was developed and validated experimentally. The current research also caters to the incorporation of failure mitigation strategies using design failure mode and effects analysis along with the experimental assessment of the performance of the developed actuator. The system has been benchmarked against an industry-developed stepper motor-driven actuator. It has shown promising results generating an actuation force of 257 N with 15 V input voltage, meeting the acceptable range for actuation operation. It further exhibits a 67% reduction in the weight of the drive mechanism, an 80% lesser component, a 32% cost reduction, and 8% energy savings and similar envelope dimensions for assembly compatibility with dampers and louvers for easy onsite deployment. The simplified design of the SMA actuator also enhances its efficiency and makes it easier to optimize and fine-tune performance parameters. The study further introduces two advanced SMA actuator configurations for extending the achievable actuation range. The first employs a multisloped bipenniform SMA wire arrangement with a curved fibre path to increase the attainable stroke through geometric reconfiguration, while the second introduces an SMA coil based actuator for high force and high stroke applications. The bio-inspired SMA-based linear actuator has applications ranging from building automation controls to lightweight actuation systems for space robotics and medical prosthesis.

## Introduction

Actuators are an integral part of building automation and control systems, playing a crucial role in the operation of heating, ventilation, and air conditioning (HVAC) systems and other automated systems. Actuators are used for converting electrical signals into mechanical motion, allowing for precise control of system components such as valves, dampers, and motors. In building automation and control applications, it plays a crucial role by enabling systems to perform specific tasks, such as opening and closing valves and adjusting airflow or temperature with an aim to optimize building performance, energy efficiency, and occupant comfort. Increasing demand for energy-efficient and sustainable building solutions [1] necessitates the development of lightweight and cost-effective actuation systems, which is accelerating the growth of actuators in building automation and control applications.

Shape Memory Alloys (SMA) are a distinct set of smart/intelligent materials capable of recovering their shape at an elevated temperature phase. The increase in SMA wire temperature under high loads can lead to shape recovery, which results in high actuation energy densities compared to different smart materials that exhibit direct coupling. When SMAs are subjected to mechanical cyclic loading, they can absorb and dissipate mechanical energy by exhibiting reversible hysteretic shape change under specific conditions. These unique characteristics have made SMAs desirable for sensing [2–4], vibration damping [5–7], and especially for actuation applications [8]. The SMA wire considered exhibits the shape-memory phenomenon that occurs in the nickel-titanium alloy family. The temperature at which such phase transformation occurs depends upon the exact composition of the alloy. In the austenite phase, which is present above the transformation temperature, the material shows high elastic modulus and cannot be easily deformed under load.

*Corresponding author:
Bishakh Bhattacharya (bishakh@iitk.ac.in)

| **Research articles** | **SMA based actuator designs** | **Force output** | **Power consumption** | **SMA wire dimension** |
|---|---|---|---|---|
| Reynearts et al. (1998) *[9]* | Shape memory alloy actuator | 21-27 N | 30 W | 360 mm |
| Mosley et al. (1999) *[10]* | SMA bundle actuator | 445 N | 126 W | 48 SMA wire bundles |
| Shin et al. (2005) *[11]* | Hydraulic linear actuator with thin film SMA | 198 N | 20V 80A | – |
| Yuen et al. (2014) *[12]* | Embedded actuator with robotic fabric | 9.6 N | 12 W (max) | 1286 mm |
| Lee et al. (2019) *[13]* | Tendon-based soft robotics actuator | 25-30 N | – | 350 mm |
| Proposed Actuator | Bio-inspired SMA based linear actuation system | 257 N | 73.5 W | 6 SMA wire stacks (900 mm each) |

Table 1 The open literature shows a comparative analysis of the existing SMA-based actuators based on the force output, power consumption, and geometrical arrangement of SMA wires. The bio-inspired shape memory alloy-based hierarchical actuator proposed in the current research generates a maximum measured force of 257 N when actuated by a power supply of 15 V pulse voltage with a current of 4.9 A, and the corresponding power consumption is 73.5 W. $'-'$ denotes that the data was not available in the reported article.

In general, shape memory alloys (SMAs) display two phases that vary with temperature: low-temperature and high-temperature phases. These phases possess distinct characteristics attributed to the existence of varying crystal structures. The low-temperature phase known as martensite ($M$) exhibits monoclinic, orthorhombic, or tetragonal crystal arrangements. On the other hand, the high-temperature phase called austenite ($A$) has a cubic crystal structure. The shear lattice distortion results in the transformation from one phase to another. The forward ($A \rightarrow M$) and reverse ($M \rightarrow A$) phase transformations form the base for the unique behavior of SMAs. Due to their characteristics properties, SMAs can provide a high level of control and precision in actuation systems [14–17], enabling them to achieve high actuation force with minimal power consumption [18,19]. In addition, SMAs can be used in a variety of environments, making them highly suitable for applications in MEMS devices [20–23], biomedical [24–27], robotic applications [28–31] and various other applications [32–35].

Conventionally, an electromechanical actuator leverages force using a motor-gear train mechanism; however, this approach has precision limitations, requires additional components, and increases assembly time and cost. In addition, the need for variable actuation force and speed outputs necessitates an array of motor and gear combinations. In contrast to conventional electromagnetic actuation systems, shape memory alloy (SMA)-based actuators offer several advantages. A SMA based actuator can accommodate various actuation outputs with minimal modification while leveraging force with high precision and response time. As a complete solid-state actuator with a high specific work-force, the SMA-based system can be miniaturized to a significant extent. Additionally, the SMA-based system operates silently, making it ideal for noiseless operations. In addition, a greater number of components necessitates additional tooling combinations, which is an expensive endeavor. A SMA wire can serve as a better alternative for components such as motors and gears, resulting in at least a 50% reduction in production costs and amortization. Overall, SMA-based actuators provide a unique combination of high energy density, compact size, low power consumption, and durability, making them suitable for a wide range of applications in robotics, aerospace, automotive, and other industries.

Table 1 shows the SMA-based actuators available in the open literature reported actuation force lesser than 30 N [9,12,13]. The parallel arrangement-based linear actuators reported the SMA wire length in the range of 350 mm; hence, the overall envelope packaging of the system exceeds 350 mm [9,13]. The bipennate-based SMA actuator utilizes 5.4 m of SMA wire for actuation; however, by virtue of the symmetrical bipenniform arrangement, the SMA wires are accommodated within the overall dimensions of 165 mm × 100 mm × 101 mm using a compression-type bias spring. As evident from Table 1, on average, the net actuation force produced by the bipennate-based SMA actuator is eight to ten times more than other SMA-based actuator designs. The force produced in a single fiber causes macro-level muscle force generation in bipennate musculature. Thus, the current system also gives the functionality of adding even more unipennate SMA branches in the stipulated length. Hence, an even higher force can be generated in the same actuator dimension.

On the other hand, while the SMA-based actuators available in the literature reported a much higher force, as shown in Table 1, they had significantly higher power consumption than the SMA-based linear actuation system with a bipenniform configuration. Upon comparison, Mosley et al. (1999) [10] had a parallel configuration of 48 SMA wire bundles, with approximately 2.6 times the SMA length, to yield a force of approximately 1.7 times the force generated by the current bipennate-based SMA actuator. Similarly, Shin et al. (2005) [11] required approximately 22 times the power supplied to yield a force closer to 0.8 times the force reported in the current research, as shown in Table 1. In addition, the bio-inspired shape memory alloy-based hierarchical actuator generates a maximum measured force of 257 N when actuated by a power supply of 15 V pulse voltage with a current of 5.6 A, and the corresponding power consumption is 84 W.

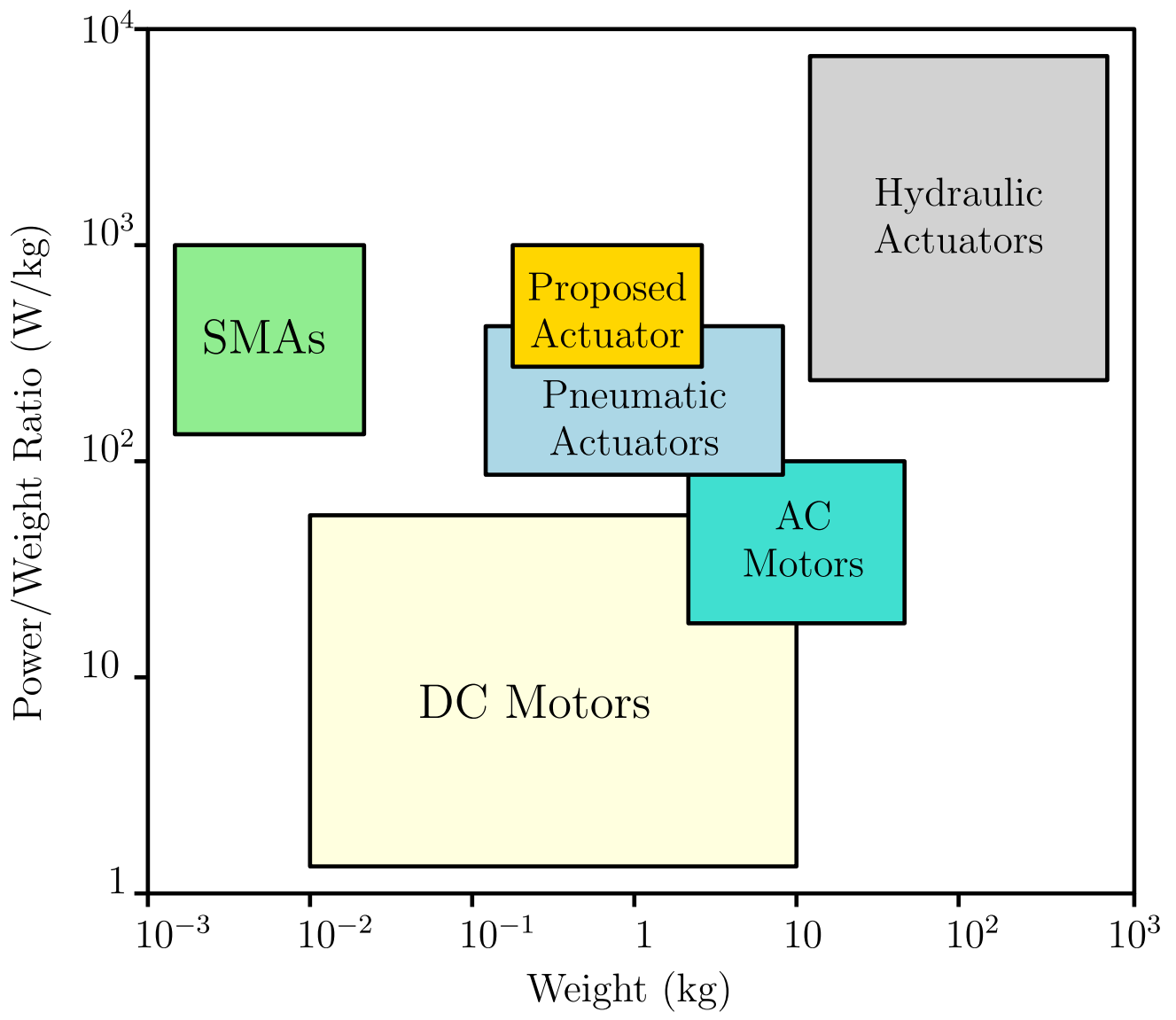

Figure 1 The distribution of the Power-to-weight ratio with the weight of different actuator categories. Schematic representation of power density as a function of weight for the most common actuator technologies. While hydraulic actuators have a high power/weight reduction. The power consumption range for DC motor-based actuation systems usually ranges from 10 - 1000 W.

Figure 1 demonstrates that shape memory alloy-based advanced systems offer a high power-to-weight ratio, allowing for a significant reduction by at least two or three orders of magnitude in actuator size while maintaining the same force-to-weight output. The current research aims to develop a new generation of actuation systems that are lighter, simpler, and more compact compared to conventional systems while still possessing the necessary strength and energy efficiency. The proposed methodology involves utilizing shape memory alloy wires arranged in a bipennate muscle configuration. The bio-inspired shape memory alloy (SMA) actuator design presented in this study has shown promising results. It exhibits a 67% reduction in the weight of the drive mechanism, an 80% decrease in the number of components, a 32% cost reduction, and 8% energy savings when compared to a conventional stepper motor-driven actuator (VA-4233 by Johnson Controls) commonly used in the HVAC building automation and controls domain.

Furthermore, the compatibility of the bio-inspired SMA actuator design with stroke amplification mechanisms has been explored. The paper discusses four distinct types of actuators based on utility parameters, achieved through different combinations of SMA elements, geometric arrangements and stroke amplification mechanisms. These categories include a high-force-low-stroke system (SMA wires without any amplification mechanism), a medium-force-medium-stroke system (SMA wires with an amplification unit), a multisloped bipenniform architecture (MSBA) developed from the single sloped bipenniform architecture (SSBA) to increase the attainable stroke through geometric rearrangement without an amplification mechanism, and a high-force-high-stroke system (SMA coils without any stroke amplification). To ensure reliability and optimize the design parameters, design failure mode and effects analysis (DFMEA) and finite element analysis were performed. These analyses helped identify potential failure modes and critical dimensions, enabling the implementation of failure mitigation strategies. The actuator model was fabricated using Markforged Metal X 3D printer with atomic diffusion additive manufacturing (ADAM) technology.

A multiphysics modelling of multi-layered bipenniform configuration-based SMA actuator was developed in this study. The mechanical behavior and performance of the linear actuation system were investigated through experimental and analytical methods. Differential scanning calorimetry (DSC) was performed on the SMA wire to capture transformation temperature data relevant to the exhibition of the shape memory effect. The performance of the actuator was assessed under various input voltage conditions, investigating force, stroke, and temperature distribution. Based on the aforementioned studies, the bioinspired SMA-based linear actuation system has shown the capability of generating a force of 257 N, falling within an acceptable range for actuation force and meeting benchmarking criteria compared to the VA-4233 actuator. The study also conducted a comparative analysis of the performance of the actuator under varying input voltage conditions and different bias spring constants. In addition, a multisloped bipenniform architecture was introduced as a geometric extension of the single sloped configuration to increase the attainable stroke within a reduced lateral envelope. Finally, a proof-of-concept for the high-force-high-stroke actuator design was developed

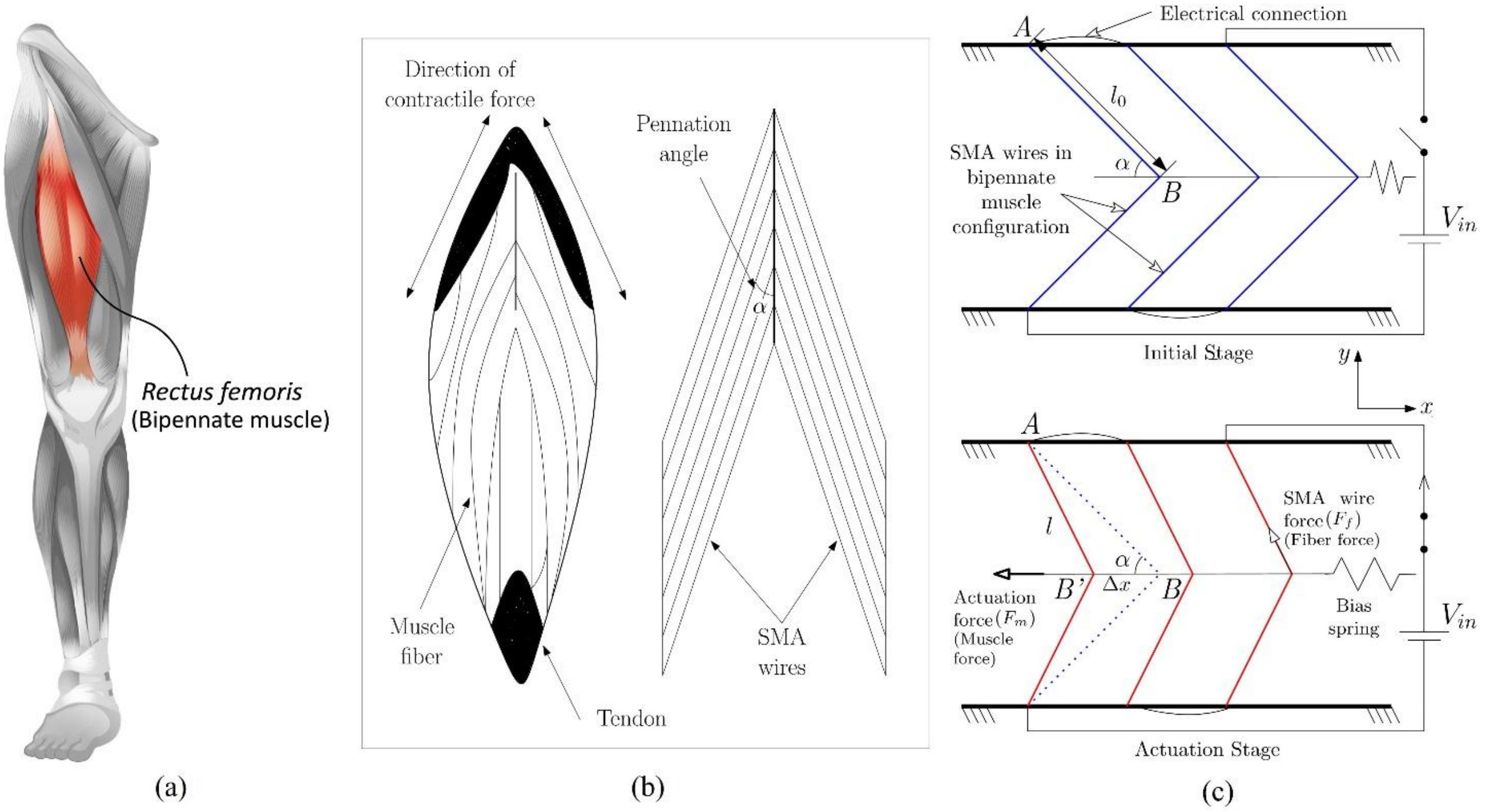

Figure 2 (a) highlights the *rectus femoris* muscle, part of the quadriceps group of the human body, whose fibers are arranged in a bipenniform manner and is situated in the middle of the front of the thigh; (b) shows the *gastrocnemius* pennate muscle tissue with the direction of contraction of muscle fibers and contractile force generation and illustrates the engineering arrangement of SMA wires analogous to muscle fibers in the bipennate muscular architecture, (c) depicts the schematic of SMA wires in a bipenniform configuration in the initial state and under actuation where the SMA wires have contracted because of the phase transformation as represented by red color lines [37].

and experimentally tested, highlighting its potential impact in constructing optimized and miniaturized actuators for applications such as space robotics and medical prosthesis.

## Methods

### *2.1. Bioinspiration: SMA-based bipenniform architecture design*

The design of muscle architecture is a critical factor in the development of bio-inspired actuators, which strive to create technologies that mimic the movement and functions of natural muscles. Actuators inspired by biological systems can be created by comprehending the basic principles of muscle design and how it relates to the operation of mammalian musculature. Such bio-inspired design is more effective and adaptable than conventional actuator technologies. Muscle architecture refers to the macroscopic configuration of muscle fibers and is broadly divided into parallel and pennate arrangements. These muscle configurations are essential to the functionality of mammalian musculature and can serve as a useful basis for the development of bio-inspired actuators.

The relatively short muscle fibers in pennate muscles are positioned at an angle to the force-generating axis. These muscles can commonly be found in mammals' body parts that work against high static loads. When a mammal, like a horse, is standing, its main task is to hold its weight against gravity, and pennate muscles are perfectly suited for this job since they can exert a lot of force relative to their size.

On the other hand, long muscle fibers aligned in the same direction as the force-generating axis produce the net work necessary to raise the mechanical energy of the center of mass when mammals accelerate or engage in rapid motions. These muscles, also referred to as parallel muscles, are frequently found in body parts that need quick and comprehensive movements. For instance, the parallel biceps *brachii* muscles in human arms are in charge of fast movements. As observed in the human *rectus femoris* muscle in the thigh, pennate muscles engage in supporting the quasi-static motion and flexion. Compared to analogous muscles in a confined physiological region, this pennate muscle can generate a considerably higher force. In a nutshell, pennate and parallel muscles have different functional properties, and understanding these differences can help significantly in the development of bio-inspired actuators. It is clear that the functional distinctions between pennate and parallel muscles, as well as the ways in which they are used in particular parts of mammalian musculature, might serve as a helpful framework for the design of bio-inspired actuators. As depicted in Figure 2, the bipennate muscles (a class of pennate muscles with axisymmetric configuration) found in the human body served as inspiration for the research since a high-force generating actuator in a quasi-static mode was envisioned [37,38].

In a pennate muscle system, the pennation angle is inversely proportional to the effective force transmitted to the tendon. When fibers operate at unequal angles, the geometry becomes inconsistent, causing displacement to be governed primarily by the shorter fibers, which also experience higher stress. For nonlinear boundary configurations, maintaining variable-length fibers introduces significant complexity in curve control, increases structural components, and complicates repair and maintenance, ultimately raising assembly cost. Therefore, for engineering simplicity and functional consistency, we adopt a linear boundary and maintain a uniform pennation angle across all fibers.

## 2.2. Mathematical Modelling

A mathematical model for a multi-layered bipenniform configuration-based SMA actuator has been developed. The set of implicit governing equations comprises the constitutive and phase transformation equations to determine the shape memory phenomenological behavior, dynamics and kinematics equations related to bipennate muscle architecture and heat transfer model for energy balance consideration.

$$\dot{\sigma} = E\dot{\epsilon} + \theta_T\dot{T} + \Omega\dot{\xi} \tag{1}$$

Equation (1) characterizes the SMA wire constitutive law governing the stress ($\sigma$) induced, dependent on the strain ($\epsilon$) developed, temperature ($T$) of the SMA wire, and martensite volume fraction ($\xi$) during transformation. The Young's modulus $E$ of SMA is phase-dependent and is governed by $E = \xi E_M + (1-\xi)E_A$ , where $E_M$ and $E_A$ represent the modulus of elasticity for the martensite and austenite phases, respectively. $\theta_T$ considers the factor for thermal expansion and the $\Omega = -E\epsilon_L$ represents the effect of phase transformation, with $\epsilon_L$ being the maximum recoverable strain in the SMA wire.

$$\dot{\xi} = \eta_\sigma\dot{\sigma} + \eta_T\dot{T} \tag{2}$$

**(Reverse transformation)**

$$\xi = \frac{\xi_M}{2}[cos[a_A(T - A_s) + b_A\sigma] + 1] \tag{3}$$

if $\dot{T} - \frac{\dot{\sigma}}{C_A} > 0$ and $\left(A_s + \frac{\sigma}{C_A}\right) \leq T \leq \left(A_f + \frac{\sigma}{C_A}\right)$

$$\eta_\sigma = \frac{\xi_M}{2}\left(\frac{a_A}{C_A}\right) sin\left[a_A\left(T - A_s - \frac{\sigma}{C_A}\right)\right]$$

$$\eta_T = -\frac{\xi_M}{2}(a_A)\, sin\left[a_A\left(T - A_s - \frac{\sigma}{C_A}\right)\right]$$

**(Forward transformation)**

$$\xi = \frac{1-\xi_A}{2} cos\left[a_M\left(T - M_f\right) + b_M\sigma\right] + \frac{1+\xi_A}{2} \tag{4}$$

else if $\dot{T} - \frac{\dot{\sigma}}{C_M} < 0$ and $\left(M_f + \frac{\sigma}{C_M}\right) \leq T \leq \left(M_s + \frac{\sigma}{C_M}\right)$

$$\eta_\sigma = \left(\frac{1-\xi_A}{2}\right)\left(\frac{a_M}{C_M}\right) sin\left[a_M\left(T - M_f - \frac{\sigma}{C_M}\right)\right]$$

$$\eta_T = -\left(\frac{1-\xi_A}{2}\right) sin\left[a_M\left(T - M_f - \frac{\sigma}{C_M}\right)\right]$$

else, $\dot{\xi} = 0$

The phenomenological model developed by Liang [39] and Brinson [40] was based on the phase kinetics and was complex to incorporate into engineering applications. Elahinia and Ahmadian [41,42] later modified the phase transformation conditions and proposed an enhanced phenomenological phase transfer model. The time dependent martensite volume fraction can be expressed as a function of stress rate and temperature gradient given by Equation (2). Equation (3) describes the transformation from martensite to austenite phase, while Equation (4) denotes the austenite to martensite phase. $\xi_M$ and $\xi_A$ are the martensite volume fraction reached before heating and cooling respectively. The curve-fitting parameters are represented by $a_A = \pi/\left(A_f - A_s\right)$, $a_M = \pi/\left(M_s - M_f\right)$, $b_A = -a_A/C_A$, $b_M = -a_M/C_M$, and $C_A$ and $C_M$, $T$ denotes the SMA wire temperature, $A_s$ and $A_f$ are austenite phase start and finish temperatures, whereas $M_s$ and $M_f$ are martensite start and finish temperatures, respectively.

$$m_{wire}c_p\dot{T} = \frac{V_{in}^2}{R_{ohm}} - A_c h_T(T - T_\infty) + m_{wire}\Delta H\dot{\xi} \tag{5}$$

Equation (5) represents heat energy balance model wherein SMA wire is being heated via applying potential difference at the end of the wires, the latent heat of transformation ($\Delta H$) determines the heat energy associated with shape memory alloy phase change. With radiation effects to be considered negligible, the heat loss in the SMA wire is characterized by forced convection and specific heat capacity term determines the heat energy stored in the SMA wire. $m_{wire}$ denotes the total mass of SMA wire deployed in the actuator, $c_p$ is the specific heat capacity of the SMA, $V_{in}$ represents the magnitugde of input voltage supply, $R_{ohm}$ is the phase-dependent resistance of SMA governed by; $R_{ohm} = (l/A_{cross})[\xi r_M + (1-\xi)r_A]$ with $r_M$ and $r_A$ as resistivity values in martensite and austenite phases, respectively. $A_c$ and $A_{cross}$ are the curved surface area and cross-sectional area of the SMA wire, respectively, whereas $T$ and $T_\infty$ represent the SMA wire and ambient temperature values, respectively.

| Parameter | Notation | Value |
|---|---|---|
| Pennation angle | $\alpha$ | 40° |
| Number of unipennate branches | $n$ | 60 |
| SMA wire diameter | $d$ | 0.51 mm |
| Initial bias spring deformation | $x_0$ | 27 mm |
| Bias spring stiffness | $K_x$ | 3.36 N/mm |
| Max. transformation strain | $\epsilon_L$ | 0.04 |
| Initial length of each unipennate wire | $l_0$ | 90 mm |
| Density of the wire | $\rho$ | 6450 kg/m$^3$ |
| Specific heat capacity | $c_p$ | 836.8 J/kg-K |
| Convective heat transfer coefficient | $h_T$ | 85 W/m$^2$-K |
| Latent heat of transformation | $\Delta H$ | 24 kJ/kg |
| Ambient temperature | $T_\infty$ | 18 °C |
| Thermal expansion factor | $\theta_T$ | 0.55 MPa/°C |
| Input voltage | $V_{in}$ | 15 V |
| Modulus of elasticity (Austenite phase) | $E_A$ | 75 GPa |
| Modulus of elasticity (Martensite phase) | $E_M$ | 28 GPa |
| Slope (Austenite phase) | $C_A$ | 10 MPa/°C |
| Slope (Martensite phase) | $C_M$ | 10 MPa/°C |
| Resistivity (Austenite phase) | $r_A$ | 100 μΩ-cm |
| Resistivity (Martensite phase) | $r_M$ | 80 μΩ-cm |
| Austenite start temperature | $A_s$ | 78.2 °C |
| Austenite finish temperature | $A_f$ | 82.1 °C |
| Martensite start temperature | $M_s$ | 54.5 °C |
| Martensite finish temperature | $M_f$ | 48.9 °C |

Table 2 The list of variable description and input conditions used for performing simulation of the analytical modelling of the multi-layered bipenniform configuration-based SMA actuator.

$$F_a = nF_{SMA} \cos\alpha - K_x(\Delta x + x_0) \tag{6}$$

$$\Delta x = \frac{n\sigma A_{cross} \cos\alpha - K_x x_0}{\frac{n}{l_0} A_{cross}\left[-E\cos^2\alpha + \frac{\sigma}{1-\epsilon}\sin^2\alpha\right]} \tag{7}$$

$$\dot{\epsilon} = \frac{(\Delta\dot{x})\cos\alpha}{\sqrt{l_0^2 - 2l_0(\Delta x)\cos\alpha}} \tag{8}$$

The actuation force is generated collectively by each unipennate SMA wire, as depicted in Figure 2(c). Equation (6) defines the relation between unipennate SMA wire force ($F_{SMA}$) and bias spring stiffness with the actuation force ($F_a$) of the bipenniform configuration. The total displacement or stroke ($\Delta x$) of the actuator as a function of stress induced and strain developed in the SMA wire is given by Equation (7). With respect to the bipennate muscle architecture design, $n$ denotes the number of unipennate branches, $\alpha$ is the angle of pennation, $F_{SMA}$ represents unipennate wire force, and $F_a$ is the resultant actuation force generated by the actuator. The stiffness of the bias spring is denoted by $K_x$ , whereas $x_0$ represents the initial deformation in the bias spring to maintain required pre-tension in the SMA wires for bipenniform configuration, and $\Delta x$ is the stroke of the actuator. Equation (8) governs the relation between strain developed in the SMA wire ($\epsilon$) and displacement ($\Delta x$) of the actuator where $l_0$ is the initial length of the unipennate, and $l$ is the length of the unipennate at any given time. Table 2 lists the parameters and its values along with the input conditions used for performing simulation of the multi-field coupled mathematical model in MathWorks Simulink R2020b environment.

## *2.3. Stroke amplification mechanism and DFMEA*

SMA actuators have many benefits over traditional actuation systems, including compact size and design simplicity. However, SMA-based actuation elements have limitations while achieving long-stroke displacements. SMA wires have a strain limit, ranging from 3 - 6%, which is still insufficient to meet the requirements of many engineering applications. Additionally, the presence of a bias compression spring in the actuator presented can limit its ability to achieve the full stroke potential, as shown in Figure 3(b). Therefore, using a stroke amplifier can help overcome these limitations and increase the overall performance of SMA-based actuators. By using a stroke amplifier, the output displacement can be increased without requiring a larger input stroke. This can allow SMA-based actuators to be used in broader applications and provide more precise and efficient motion.

The current research presents a modular actuator design compatible with both compliant stroke amplification mechanism (CSAM) and hydraulic stroke amplification mechanism (HSAM), each of which has its own benefits. CSAMs have numerous advantages over conventional rigid mechanisms, such as greater precision and accuracy, compactness, lightweight, reduced cost, and better design flexibility. HSAMs, on the other hand, also offer several benefits, including high stroke output, high power density, seamless and precise motion, adjustability, low maintenance, and safety, making them suitable for various heavy-duty lifting and industrial automation applications. Hence, the current research focuses on designing modular actuators to incorporate the advantages HSAMs offer into the overall system, as illustrated in Figure 3(c). To achieve this, an axisymmetric finite element (FE) analysis was performed on the HSAM prototype. Figure 3(d) depicts the meshing of the prototype, illustrating the division of the model into discrete elements. The FE analysis incorporated a total of 2574 elements, comprising 2492 triangular, 262 quadrilateral, 196 edges, and 19 vertex elements. A fluid-structure interaction (FSI) physics study has been carried out where the stroke amplification factor of 11.8 was obtained. The initial and final

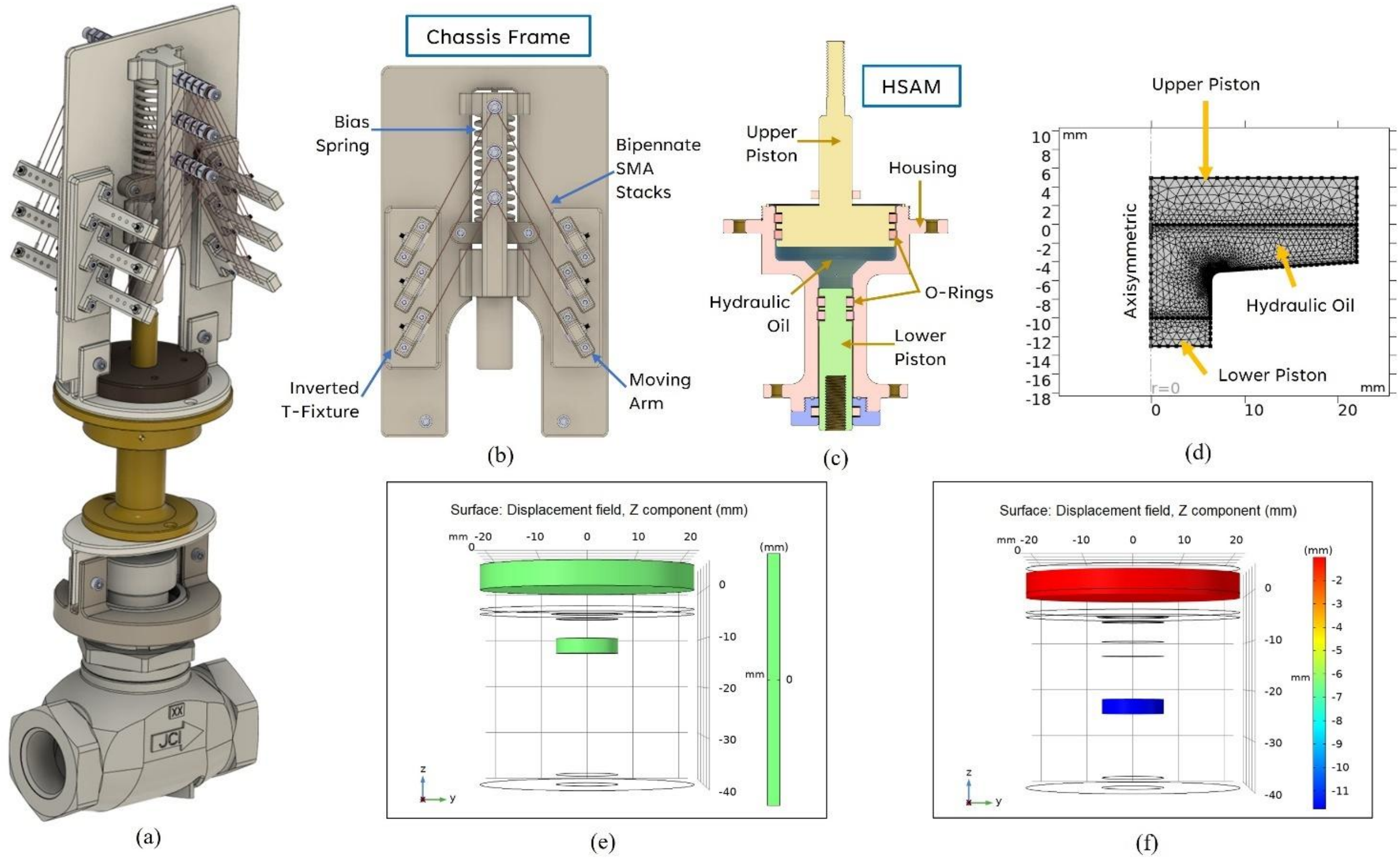

Figure 3 (a) showcases the CAD model assembly of a bioinspired SMA-based linear actuator integrated with a hydraulic stroke amplification mechanism (HSAM) and coupled with a globe valve. (b) depicts the components mounted on the actuator chassis frame, such as a bias compression bias spring, inverted T-fixture, moving arm, and SMA wire stacks arranged in a bipenniform configuration. (c) displays a section view of HSAM highlighting the positions of the upper and lower pistons, O-rings, and hydraulic oil. (d), (e) and (f) illustrate the snapshots of the finite element analysis conducted on the HSAM. (d) The study adopts an axisymmetric approach and illustrates the meshed geometry, with the corners containing a higher number of mesh elements. (e) The initial condition depicts both the lower and upper pistons at zero displacements. (f) shows the result wherein a 1 mm displacement of the upper piston resulted in an 11.8 mm movement of the lower piston, corresponding to a stroke amplification factor of 11.8.

conditions of the finite element analysis (FEA) conducted on COMSOL Multiphysics 6.0 are depicted in Figure 3(e) and Figure 3(f), respectively. The former represents the initial condition, while the latter represents the final condition. In the final condition, a movement of 1 mm in the upper piston led to a displacement of 11.8 mm in the lower piston.

The Design failure mode and effects analysis (DFMEA) provides a systematic framework for identifying potential risks in a new or modified product design. Initially, the DFMEA helps in identifying critical components, their functions, and potential failure modes, followed by the causes and effects of these failures. The Risk Priority Number (RPN) methodology is then utilized to analyze the associated risks and prioritize improvement efforts. The RPN represents relative risk ranking and is calculated by multiplying the scores of three factors: Severity, Occurrence, and Detection. *Severity* assesses the impact of the failure mode, with 1 representing the least safety concern and 10 indicating the highest safety concern. *Occurrence* represents the likelihood of a failure occurring, with 1 indicating the lowest occurrence and 10 indicating the highest. *Detection* assesses the probability of detecting a failure, with 1 being the highest chance of detection and 10 indicating the lowest.

In the case of the bio-inspired shape memory alloy-based linear actuator system, the DFMEA has been carried out to identify potential failures. Table 3 displays the results of the design FMEA analysis, listing the system components such as SMA wires, 3D printed parts, connecting crimps, conducting wires, bias spring, and hydraulic oil, along with their roles in the overall actuation system. The critical items are ranked in order of priority based on their calculated Risk Priority Numbers (RPNs). DFMEA serves as an effective quality-control tool, enabling the incorporation of mitigation strategies during the design stage. For instance, to address potential failures related to 3D-printed support fixtures used for holding SMA wires, a high glass transition temperature filament, such as poly-carbonate, has been selected for fabrication. Similarly, silicone o-rings have been placed at appropriate interfaces between the piston and cylinder to prevent hydraulic oil leakage.

| Item | Function | Causes | Potential failure mode | Potential effects of failure | Severity (S) | Occurrence (O) | Detection (D) | Risk Priority Number (S×O×D) |
|---|---|---|---|---|---|---|---|---|
| SMA wire | Responsible for actuation | Overheating | Partial failure | Lesser force output | 7 | 8 | 9 | 504 |
| 3D printed parts | Support fixtures | Melting due to heat accumulatio n near SMA | Partial failure | Stroke & force output will be affected | 6 | 5 | 7 | 210 |
| Hydraulic oil | Pressure transmission across pistons | Oil leakage | Partial failure | Lesser stroke output and high stress on housing | 8 | 4 | 6 | 192 |
| Crimps | Responsible for SMA attachment | Breakage due to over-tightening | Full/ Partial failure | Complete shut off or lesser force output | 8 | 3 | 7 | 168 |
| Electrical wires | Responsible for electical connectivity | Soldering breakage | Full/ Partial failure | No/less force and power consumption | 9 | 3 | 6 | 162 |
| Bias spring | Providing tension & bias force in SMA architecture | Fixture break or misalign-ment | Degrade d failure | No force but continued power consumption | 7 | 3 | 5 | 105 |

Table 3 Design failure mode and effects analysis (DFMEA) for bio-inspired shape memory alloy-based linear actuator system. The system components such as SMA wire, 3D printed parts, connecting crimps, conducting wires, bias spring, and hydraulic oil have been listed alongside their role in the overall actuation system. The critical items are ranked in the order of priority based on the calculated risk priority number (RPN).

Furthermore, a static structural analysis was performed using ANSYS 2021 R2 to identify the critical areas of the HSAM housing experiencing high stress during a hydraulic oil leakage failure case. Figure 4(a) illustrates the boundary conditions in the FEA wherein the fixed constraints are applied on the top and bottom coupler yoke where the bolts are used for mounting with the chassis frame. Bolted connections are modeled with the help of joint connectors in simulation, and frictional contacts are defined at all other contacting surfaces. In the worst-case scenario, when the oil would be completely drained out due to leakage, the upper piston would exhibit a force of 257 N (maximum SMA actuation force obtained during experimentation) on the cylinder face in the actuation direction, as highlighted with red in the Figure 4(a). The meshing of the HSAM assembly consists of 1,114,596 quadratic 3-D 10-Node tetrahedral structural elements and 1,630,261 nodes, as shown in Figure 4(b). This refined meshing approach was deployed to ensure higher accuracy in the simulation results. Figure 4(c) depicts the *von Mises* stress distribution across the housing of the HSAM. The higher *von Mises* stresses are observed in the hydraulic cylinder at fillet locations which are identified as the critical areas of the housing cylinder. The yield strength of the Brass material used in the cylinder is 234 MPa and the maximum stress reported at critical locations is 5.5 Mpa which is within the design limit of the material; hence the system would safely work without any plastic yielding.

## 2.4. *Prototyping and fabrication*

Based on the bio-inspired design concept of bipennate muscle configuration, a CAD model of the SMA-powered prototype has been developed, as depicted in Figure 3(a) [43,44]. The fabrication of the model was accomplished using an additive manufacturing technique known as the fused deposition modelling (FDM) 3D printing method. FDM is the most suitable and widely utilized technique among 3D printing technologies due to its high accuracy, cost-effectiveness, and wide range of material options for creating complex geometrical parts. In FDM, components are built layer-by-layer by selectively depositing melted material filament along a predetermined path derived from a digital design file. To evaluate the geometric dimensioning and tolerancing (GD&T), a proof-of-concept model was fabricated using white polylactic acid (PLA) filament with the help of Ultimaker 3 Extended 3D printer, as shown in Figure 5(g).

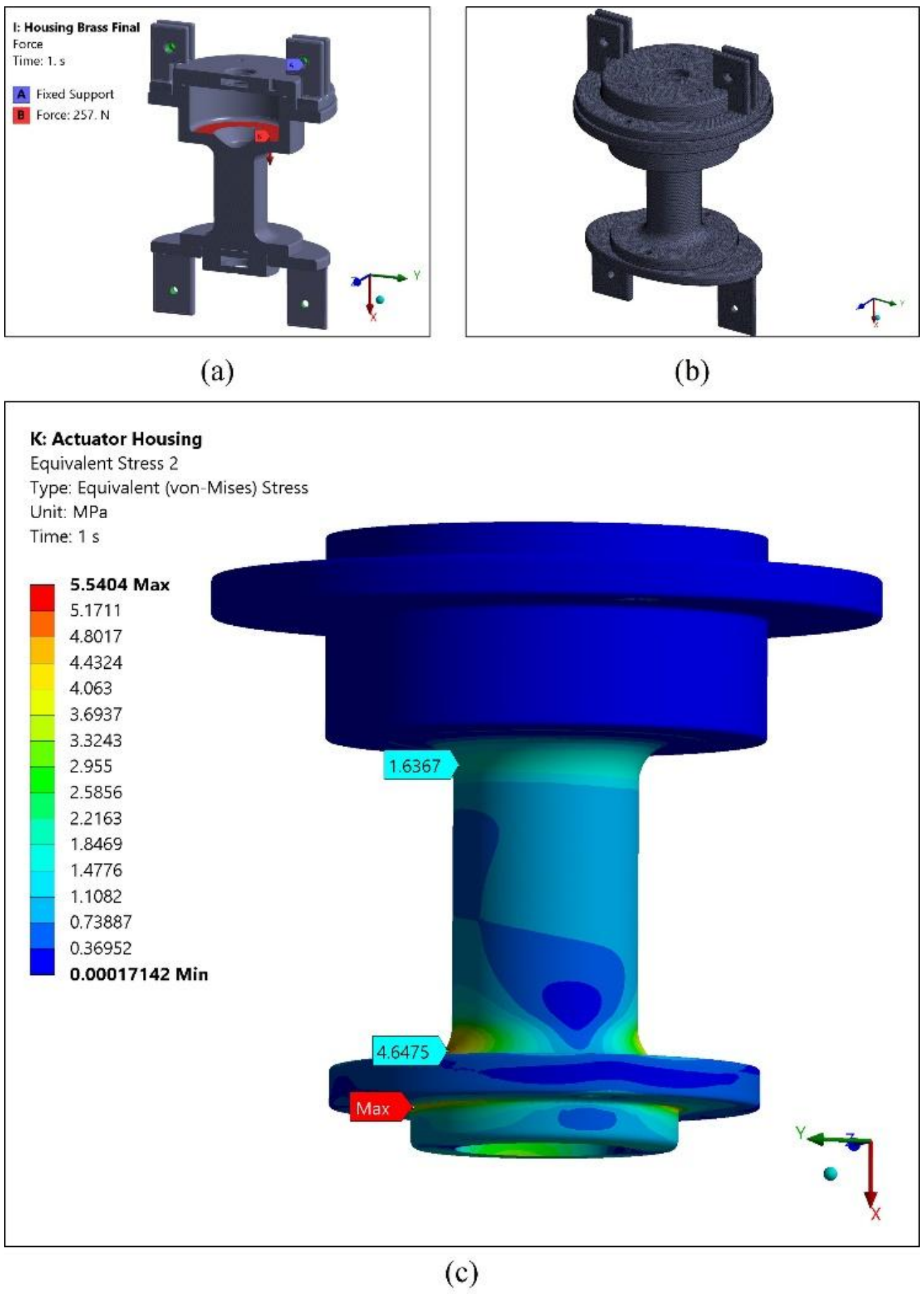

Figure 4 Static structural analysis was carried out on HSAM housing using ANSYS 2021 R2 during a hydraulic oil leakage failure case. (a) illustrates the boundary conditions applied for performing FEA, and (b) depicts the meshing having quadratic 3-D 10-Node tetrahedral structural elements and nodes. (c) highlights the von Mises stress distribution across the housing of the HSAM, wherein the maximum stress reported at the fillet region is 5.5 MPa which is within the design limit (Yield strength of brass = 234 MPa) of the material.

The components of the final model are manufactured using a Markforged Metal X machine based on atomicdiffusion additive manufacturing (ADAM) technology, utilizing 17-4 PH stainless steel as the raw material, as depicted in Figure 5(b). The SMA wires (0.51 mm diameter), specifically *Flexinol* actuator wires from *Dynalloy, Inc.*, are arranged in a bipennate muscle configuration with a vertical stack. The two ends of the wires are fixed on the inverted T-fixture and pivoted on the bolt of the moving arm. One vertical stack utilizes 1 m of SMA wire wounded at five levels maintaining an optimum gap of 3 mm between two subsequent SMA wires winding to avoid short-circuiting and facilitating effective convective heat loss during the cooling cycle. To accommodate more bipennate units in a compact size, the entire assembly is mirrored about the chasis frame, as as illustrated in Figure 5(e).

An electric insulation coating (Insulect SK-03) has been applied to the inverted T-fixtures to prevent electrical short-circuiting between the metal supports and SMA wires, as shown in Figure 5(f). This coating offers a di-electric strength of 59 kV/mm with stability up to 200 °C. Additionally, glass fiber sleeves are inserted into the pinholes of the inverted T-fixtures and bolts of the moving arm, providing further short-circuit protection to the SMA wire windings. The entire support assembly is sandwiched between a 3D printed poly-carbonate cover as the material is tough and has a high glass transition temperature (110–113 °C), making it suitable heat-resistant material compatible with SMA wires having high transformation temperature, as shown in Figure 5(c). The hydraulic stroke amplification mechanism (HSAM) sub-system components are manufactured using conventional machining techniques with brass as the raw material. This sub-system comprises a cylinder housing, upper and lower pistons, silicone O-rings, lock nuts, mounting couplers, and ISO-62 hydraulic oil. The assembly of the HSAM sub-system is depicted in Figure 5(d), while the complete SMA bipennate linear actuator integrated with the VG7241 Globe valve is shown in Figure 5(a).

## Experimentation

### *3.1. Differential scanning calorimetry*

As stated earlier, the shape memory effect results from the transformation of the martensite phase to austenite. Superelasticity, on the other hand, is characterized by the restoration of strain under isothermal conditions during mechanical loading and unloading cycle, which occurs when the transformation temperature threshold of the SMA wire is exceeded [43]. The solid-solid transformations in shape memory alloys (SMAs) are completely reversible and do not involve plastic deformation. Consequently, it is essential to acquire a fundamental understanding of the thermomechanical behaviour of SMA material. Differential scanning calorimetry (DSC) has been used to characterise the transformation temperatures of SMAs as illustrated in Figure 6. SMAs consist of two main phases, austenite (the high-temperature, zero-stress phase) and martensite (the low-temperature, zero-stress phase), as well as multiple transformation temperatures, including the austenite start temperature ($A_s$), austenite finish temperature ($A_f$), martensite start temperature ($M_s$), and martensite finish temperature ($M_f$). The SMA element temperature affects the actuation behaviour of SMAs, highlighting the critical importance of thermal analysis throughout the entire process. Therefore, the SMA wire sample was subjected to DSC analysis prior to experimentation to obtain phase transformation temperature data. Ensuring that the SMA wires do not exceed a safe operating temperature limit is essential, as overheating can result in dislocation movement throughout its lattice structure and degrade memory recovery.

Figure 6(a) shows the experimental setup of differential scanning calorimetry wherein a small sample (9.7 mg of SMA wire with diameter 0.51 mm) was inserted into the PerkinElmer-made DSC-8500 instrument under no-stress condition for the experimental analysis. The sample was subsequently scanned at a constant rate of 10 °C/min while the heat flow rate was monitored. Notably, the transformation temperature can be readily modified by the manufacturer and could range from cryogenic to temperatures exceeding 100 °C. Consequently, it is essential to investigate the temperature spectrum from extremely low to high temperatures range. Therefore, liquid nitrogen ($LN_2$) cooling is required to witness the transformations at reduced temperatures. The investigation required subjecting the SMA wire sample to multiple successive heating and cooling cycles. In the present study, ten heating and cooling cycles have been repeated on the sample. As shown in Figure 6(b), the rates of heating and cooling were set to 10 °C/min having a one-minute dwell period, with a maximum and minimum temperature of 130 °C and 0 °C, respectively.

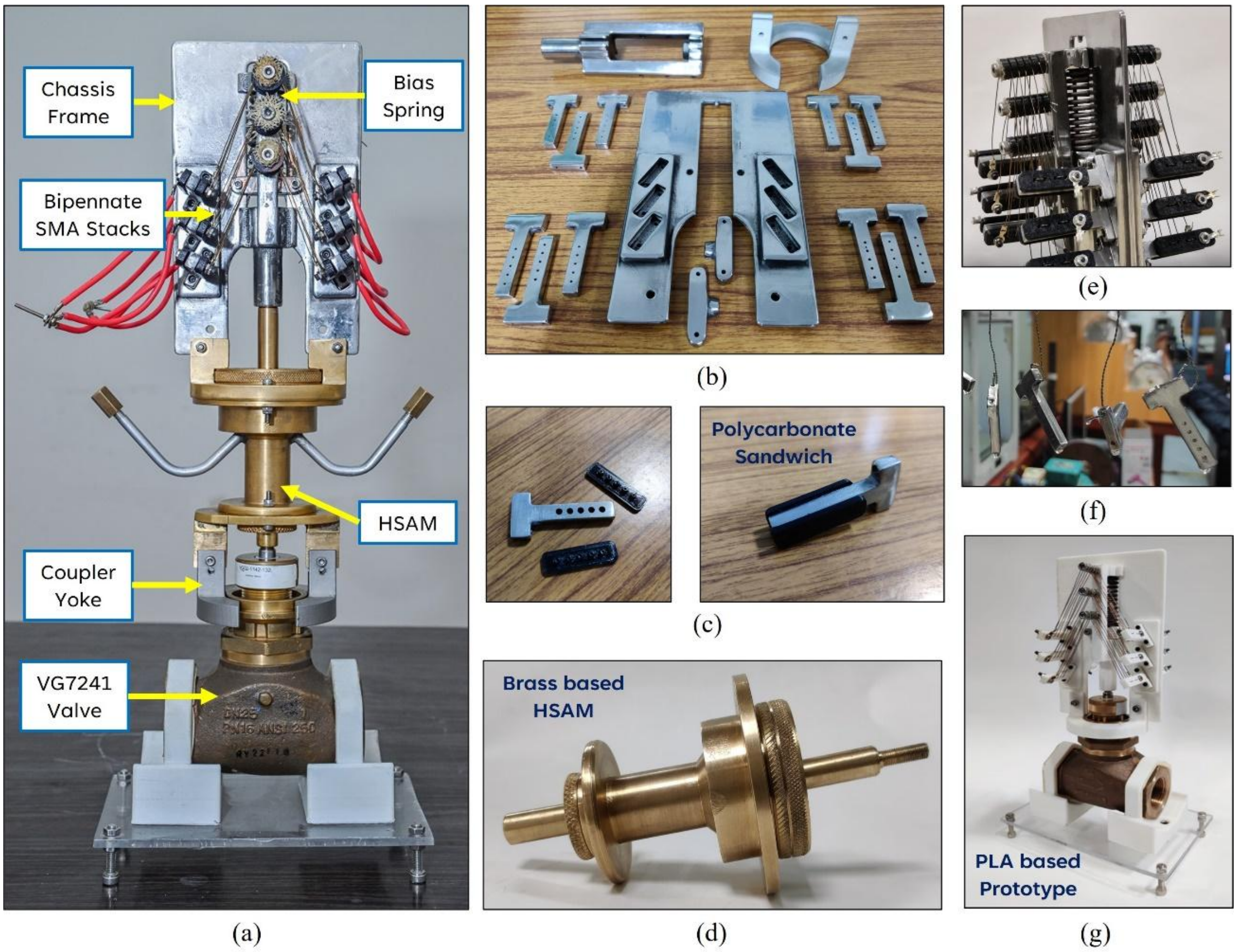

Figure 5 This figure showcases the complete actuator model assembly fabricated using additive manufacturing technology. (a) The medium force-medium stroke actuator design model, equipped with the VG7241 globe valve, consists of a metallic chassis frame and bipennate SMA wire stacks. (b) depicts the components of the actuator manufactured using a Markforged Metal X machine based on atomic diffusion additive manufacturing (ADAM) technology, utilizing 17-4 PH stainless steel as the raw material. (c) shows the metallic inverted T-fixtures incorporate a polycarbonate sandwich insulation cover chosen for its heat-resilient properties to prevent short-circuiting. (d) shows the assembled view of the hydraulic stroke amplifier manufactured using conventional machining techniques with brass as the raw material. This sub-system comprises a cylinder housing, upper and lower pistons, silicone O-rings, lock nuts, and ISO-62 hydraulic oil. (e) shows the bipenniform arrangement of SMA wires wherein one side of the central chassis frame is equipped with three on-the-plane SMA wire stacks and five stacks of SMA wires into the plane. The entire architecture is then mirrored on the opposite side of the chassis frame to ensure symmetry about the central frame. (f) shows that Insulect SK-03 insulation coating has been applied to the inverted T-fixture and is being allowed to dry. (g) displays an isometric view of the proof-of-concept model fabricated using polylactic acid (PLA) filament.

### 3.2. *Actuation force measurement*

The developed bipennate SMA actuator has a modular design so that it can be configured as either a high-force, low-stroke actuator or a medium-force, medium-stroke actuator, based on the application. One of the essential parts of this adaptability is incorporating a hydraulic stroke amplification mechanism (HSAM) with a practical stroke amplification factor of twelve. This factor signifies that the HSAM can enhance the stroke output by a factor of twelve; however, at the same time, it also reduces the force output. Hence, to evaluate the performance of the actuator, two distinct experiments have been performed. By comparing the results of both experiments, the performance of the actuator with and without the HSAM was assessed. A series of experiments were conducted to measure the relevant parameters, such as the actuator force output and stroke/displacement. Overall, the modular design framework of the actuator and its compatibility in terms of integration

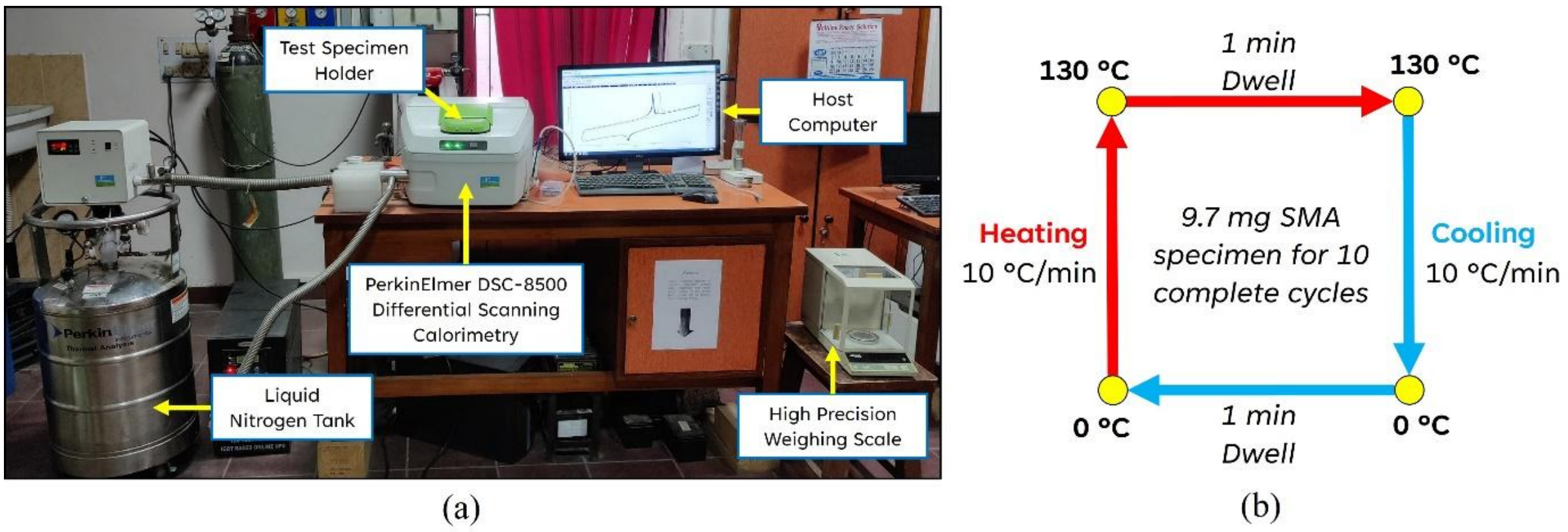

Figure 6 (a) The experimental setup for measuring thermal parameters of the SMA wire via Differential Scanning Calorimetry (DSC). A PerkinElmer DSC-8500 was used to measure the heat released or absorbed. A high precision weighing scale was utilized to accurately measure the weight of the specimen along with a liquid nitrogen tank to bring the temperature of the wire to very low temperatures (up to 0 °C). The data generated by the DSC setup is transferred to the host computer for data analysis. (b) Schematic of the DSC experiment for the 9.7 mg of SMA wire (0.51 mm diameter) sample for 10 cycles. The maximum temperature of the wire was subjected to 130 °C and the minimum 0 °C, with both the heating and cooling cycle being 10 °C/min with a 1 min dwelling period while switching between heating and cooling cycle.

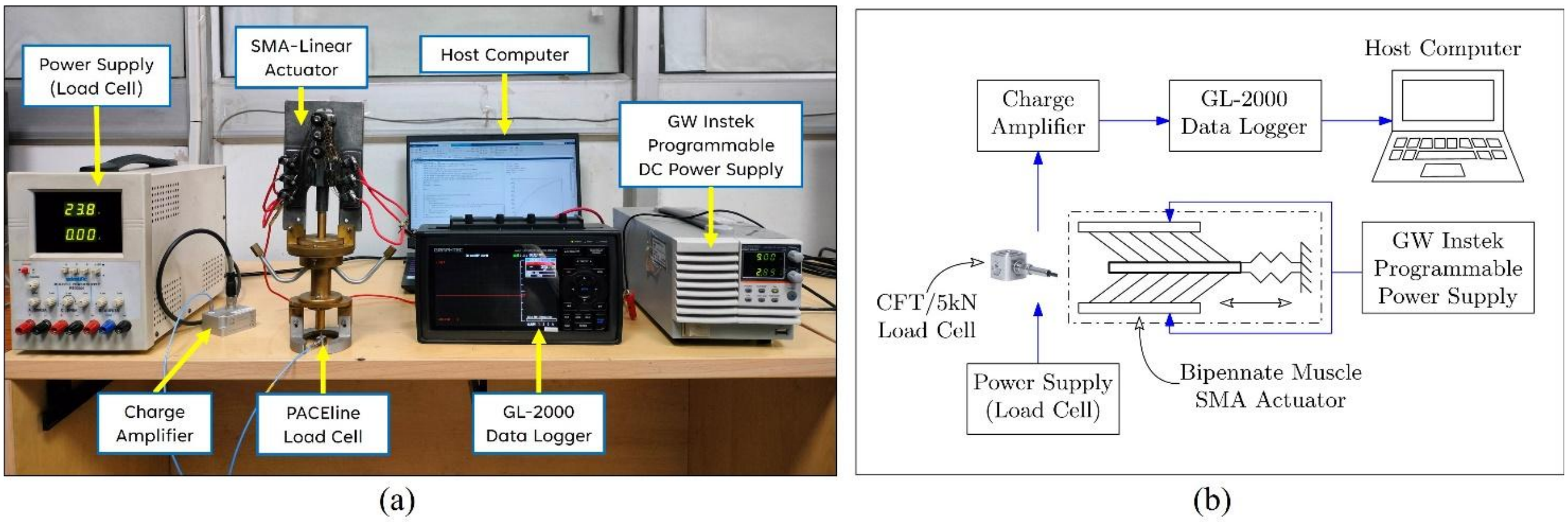

Figure 7 (a) shows the experimental setup involved in the measurement of the output force generated by the SMA-based linear actuator. A PACEline CFT/5kN load cell, powered by a constant 24V power supply, was deployed to measure the blocked force. A GW-Instek programmable DC power source (Voltage range: 0 - 80V; Current range: 0 - 27A and Power rating: 720 W) was used to supply the input voltage pulse to the actuator. Under the effect of the actuation of SMA wire stacks, the lower piston of the HSAM comes in contact with the load cell, generating the blocked force. The load cell data is recorded using the GL-2000 data logger and subsequently transferred to the host computer for data analysis. (b) illustrates a schematic of the circuit configuration and types of equipment involved in the force measurement experiment.

with HSAM makes it advantageous and a potential alternative for a wide range of actuation applications.

As depicted in the model actuator, Figure 5(e), one side of the central chassis frame is equipped with three on-the-plane SMA wire stacks and five stacks of SMA wires into the plane. The entire architecture is then mirrored on the opposite side of the chassis frame to ensure symmetry about the central frame. As illustrated in Figure 7, a load cell is affixed to the central frame plane to measure actuation force. The SMA wires are heated using the Joule heating effect, which is facilitated by a programmable DC power supply that provides a potential differential across the SMA wires, as shown in Figure 7(a). When an input voltage is applied, and SMA wires are allowed to heat above their austenite phase transition temperature ($A_s = 78.2$ °C), it results in the recovery of its original shape, leading to the generation of actuation force. This blocked force is then measured by a piezoelectric load cell (PACEline CFT/5kN), powered by a constant DC power supply, and the voltage signal corresponding to the blocked force is recorded using a Graphtec GL-2000 data logger. Figure 7(b) shows the schematic of the types of equipment involved in the force measurement experiment. The charge amplifier is used to boost the signal from the load cell, which is then stored in the host computer via a data logger for post-processing and further analysis. The actuation force data is obtained from the corresponding voltage signal by using the conversion factor, which is a function of sensitivity and other technical parameters of the load cell as listed in Table 4.

| PACEline CFT/5kN Load Cell | | FLIR A700 Thermal Imaging Camera | | optoNCDT 1420 Laser Sensor | |
|---|---|---|---|---|---|
| *Parameters* | *Value* | *Parameters* | *Value* | *Parameters* | *Value* |
| Sensitivity | - 8.097 pC/N | Resolution | 640 × 480 | Measuring range | 50 mm |
| Nominal force | 5 kN | Field of view | 14° × 10° | Measuring rate | 4 kHz |
| Natural frequency | 40 kHz | Frame rate | 30 Hz | Repeatability | 2 μm |
| Output span | ± 10 V | Lens type | IR, $f$ = 29 mm | Linearity | ≤ ± 40 μm |
| Breaking force | 10 kN | Spectral range | 7.5 − 14.0 μm | Supply voltage | 11- 30 VDC |

Table 4 The table depicts the technical parameters of three different sensing devices used in the current experimental study: the PACEline CFT/5kN piezoelectric force transducer, the FLIR A700 high-resolution science grade thermal imaging camera, and the optoNCDT 1420 smart laser triangulation displacement sensor.

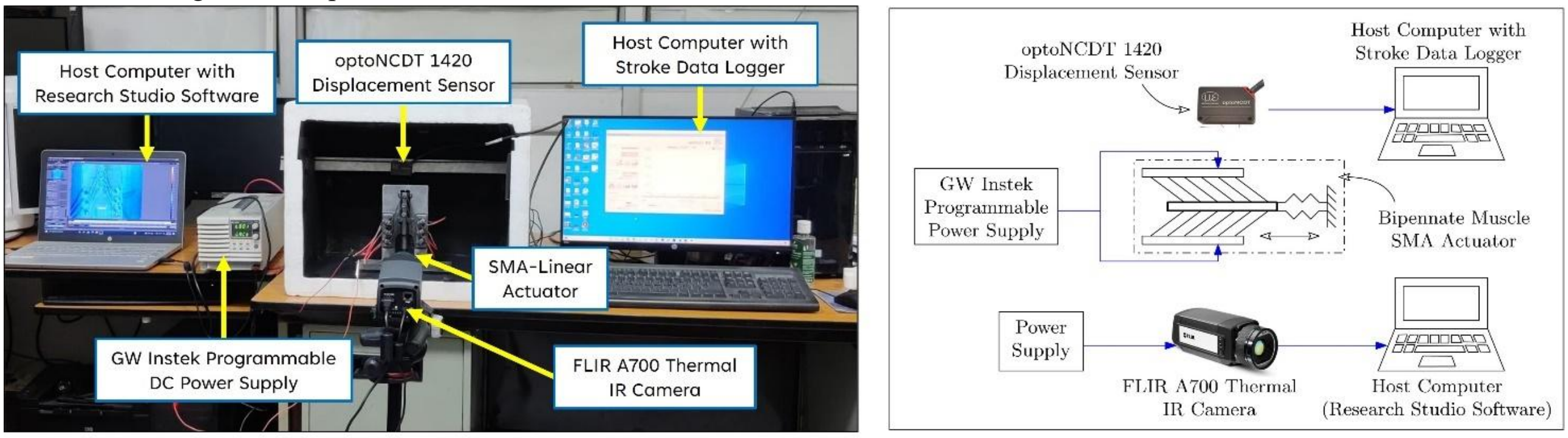

(a) (b)

Figure 8 (a) illustrates the experimental setup for capturing the stroke of the SMA-based linear actuator and monitoring SMA wire temperature using thermal imaging. An optoNCDT 1420 laser displacement sensor was used to measure the developed stroke of the actuator powered by a constant 24V DC power supply. A GW-Instek programmable DC power source (Voltage range: 0 - 80V; Current range: 0 - 27A and Power rating: 720 W) was used to supply an input pulse of 15V to the actuator. The resistive heating process increased the temperature of the wire, which was captured by the FLIR A700 thermal IR camera. The data generated by the displacement sensor and thermal IR camera were transferred to the host computers for post-processing and analysis. (b) depicts the schematic diagram of the experimental setup and circuit configuration for measuring the stroke and thermal monitoring of the system.

### 3.3. *Actuation stroke measurement and thermal imaging experiments*

The actuator stroke measurement and thermal analysis play a vital role in understanding the behavior and assessing the performance of the SMA-based actuator. Measuring displacement data is essential for accurate positioning, system calibration, and feedback control of the actuator. It enables efficient and reliable actuation operation in various industrial applications, including robotics, automation, aerospace, and manufacturing. Figure 8(a) depicts the experimental setup wherein an optoNCDT 1420 laser displacement sensor was deployed to measure the stroke generated by the actuator. When the SMA wires are actuated, they undergo contraction, causing the movable arm of the actuator to descend. This movable arm is subsequently linked to the upper piston of the HSAM, which transfers the amplified motion to the lower piston of the HSAM. The displacement of the lower piston is then accurately measured by a laser displacement sensor, allowing for precise monitoring and analysis. The displacement sensor was powered by a DC power supply, and the data of the sensor was recorded on a host computer for data post-processing, as illustrated in Figure 8(b).

During the experiment, the SMA-based actuator was also subjected to thermal monitoring, as thermal imaging can effectively understand the temperature-dependent actuation of SMA wires. To ensure accurate thermal imaging of the SMA wires, the actuator was placed inside an expanded polystyrene (EPS) box and covered with black paper to ensure a dark environment, as shown in Figure 8(a). The EPS box was used as insulation because it is a well-known insulator, effectively reducing heat transmission between the SMA-based actuator and the surrounding environment. By minimizing heat transfer, the internal temperature of the box remained stable, allowing for more noise-free measurements of the SMA wire temperature. Using a black background aided in absorbing radiation and reducing thermal signature interference. Figure 8(b) shows the schematic of the experimental setup wherein a high-resolution science-grade LWIR camera (FLIR A700) was used to monitor the temporal variation of the temperature when the SMA-based actuator was subjected to an input voltage from the programmable DC power supply. Table 4 lists the technical parameter corresponding to optoNCDT 1420 laser displacement sensor FLIR A700 LWIR camera. The post-processing and detailed analysis of the thermal imaging data was carried out using *FLIR Research Studio* software on the host computer.

## Results and Discussion

### 4.1. *Material characterization - Differential scanning calorimetry*

At the design stage of SMA-based actuation systems, it is essential to have a fundamental understanding of the various transformation phases, including austenite start temperature ($A_s$), austenite finish temperature ($A_f$), martensite start temperature ($M_s$), and martensite finish temperature ($M_f$). As depicted in the Figure 9, the heat flow rate at each temperature is determined (with the least count of 0.01 °C) using DSC, with the thermogram exhibiting the heat flow rate (in mW) on the ordinate and the temperature (in °C) on the abscissa. The thermogram displays two peaks, one signifying the endothermic transformation to austenite and the other the exothermic transformation to martensite. These peaks are characterized by a $C_0$ function that exhibits a sudden increase (or decrease), reaches a maximum (or minimum), and then decreases (or increases) before rejoining the baseline. Thus the transformation temperature is obtained by the intersection point of the tangent constructed to the steepest side of the curve and the extension of the baseline curve. Using this method, transformation temperatures were calculated, and it was observed that the steepest point on the curves had an almost parallel slope to the ordinate axis. As shown in Figure 9, the austenite transformation temperatures of the first cycle ($A_{s0} = 85.6$ °C and $A_{f0} = 88.9$ °C) differed from those of subsequent cycles ($A_s = 78.2$ °C and $A_f = 82.1$ °C), which may be a result of residual pre-strain in the wire [44]. The martensite transformation temperatures remained constant across all cycles, with $M_s$ being equal to 48.9 °C and $M_f$ being equal to 54.5 °C.

### 4.2. *Actuator performance and optimization*

The performance measurement of the actuator provides a platform for the optimization of design parameters as well as input parameters, as presented in Figure 10. According to the results of the previous investigations, it is clear that the input voltage plays a crucial role in ensuring the operation of SMA actuators. This is due to the fact that the regulation of the input voltage has a direct effect on the heating of the wires. Therefore, it was considered necessary to visualize the effect of the input voltage on the SMA wires in order to obtain a better understanding of their behavior while also trying to prevent any unfavorable outcomes that could result from overheating. To prevent overheating, the actuator force was additionally monitored using thermal imaging. It is crucial to note that thermal monitoring of the actuator is necessary to prevent the wire from exceeding its safe operational limit of 120°C (Reference: *Flexinol* wires technical datasheet provided by *Dynalloy Inc.*). This is particularly essential when the actuator is subjected to sustained voltage, which could cause irreversible damage to the SMA wires and ultimately compromise the efficacy of the system. By implementing thermal imaging into the experimentation procedure, valuable insights into the thermal behavior of the SMA actuator are obtained, thereby promoting its safe and optimal operation.

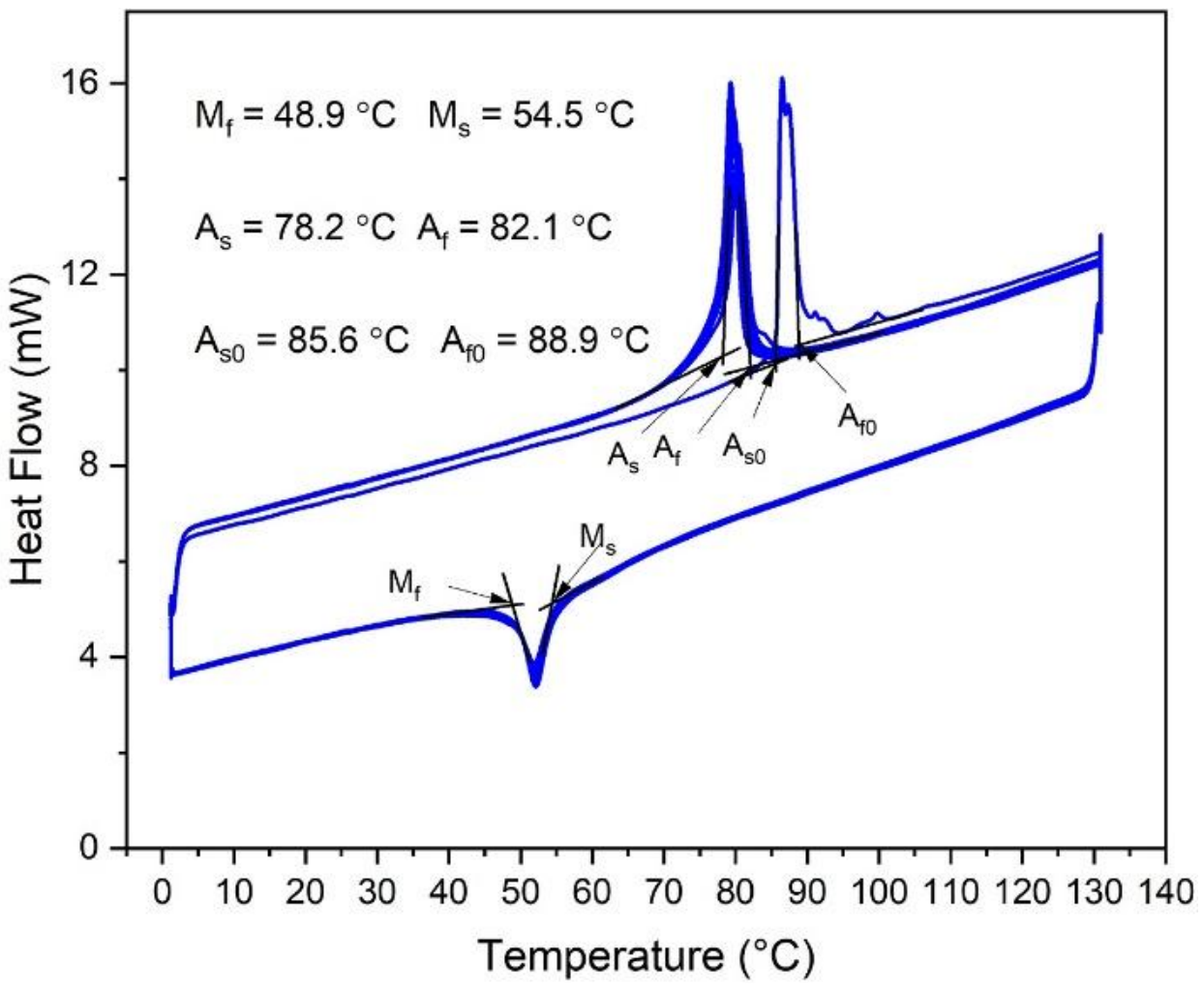

(a)

| Phase Parameter | Value |
|---|---|
| $M_f$ − Martensite finish temperature | 48.9 °C |
| $M_s$ − Martensite start temperature | 54.5 °C |
| $A_{s0}$ - First cycle austenite start temperature | 85.6 °C |
| $A_{f0}$ - First cycle austenite finish temperature | 88.9 °C |
| $A_s$ − Austenite start temperature | 78.2 °C |
| $A_f$ − Austenite finish temperature | 82.1 °C |

(b)

Figure 9 The plot depicts the experimental results of the phase transition temperatures of the SMA wire for ten cycles. The SMA sample (*Flexino*l actuator wire from *Dynalloy, Inc.*) with a 0.51 mm wire diameter weighing 9.7 mg was used in the experiment. (a) The graph displays the heat flow during the heating and cooling process of the SMA wire at a constant rate of 10 °C /min. The thermogram shows two transformations: an endothermic transformation to the austenite phase and an exothermic transformation to the martensite phase. The transformation to austenite in the first cycle exhibited a slight variation, which can be attributed to residual strain. (b) the table lists the phase transition temperature data of SMA from the DSC experiment.

In the current experiment, input voltage was varied in order to determine its effect on the generated force and the temperature of the wires. The resulting graph in Figure 10(a) demonstrates a clear positive correlation between actuation voltage and the actuation force, with an increase in actuation voltage resulting in a higher magnitude of force. In the case of 10 seconds of actuation time, a 34% increase in actuation force was observed on increasing the voltage input from 13 V (113 N) to 14 V (170 N), and an 18% increase in actuation force was observed on increasing the voltage input from 14 V (170 N) to 15 V (200 N). It is essential to note that the percentage increase in generated force decreased as the voltage was increased. It was also observed that as the voltage was increased, the increase in SMA wire temperature was captured. This behavior is consistent with established physical laws governing variations in temperature as a function of the Joule heating effect, convective heat loss, and latent heat of

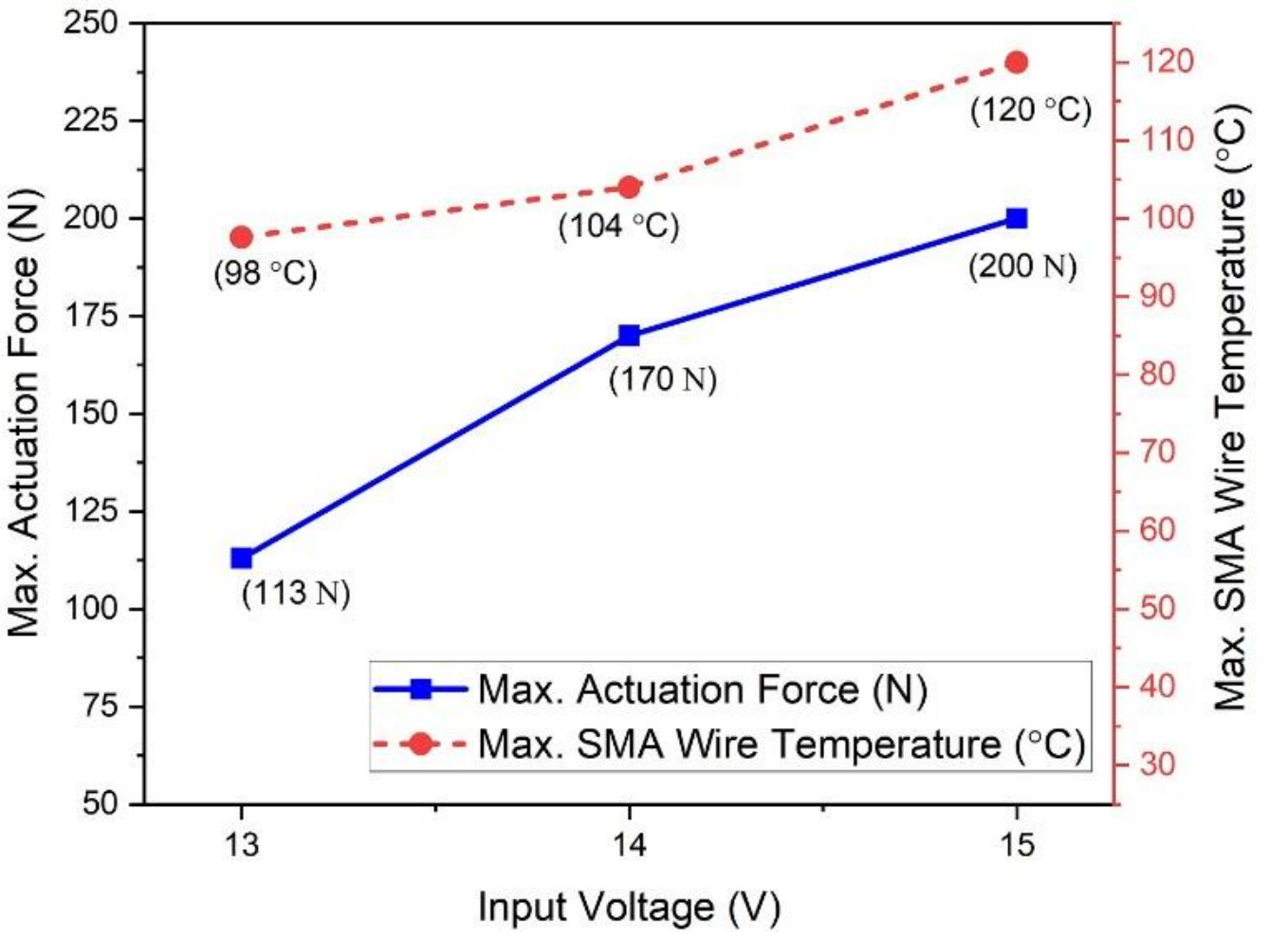

(a)

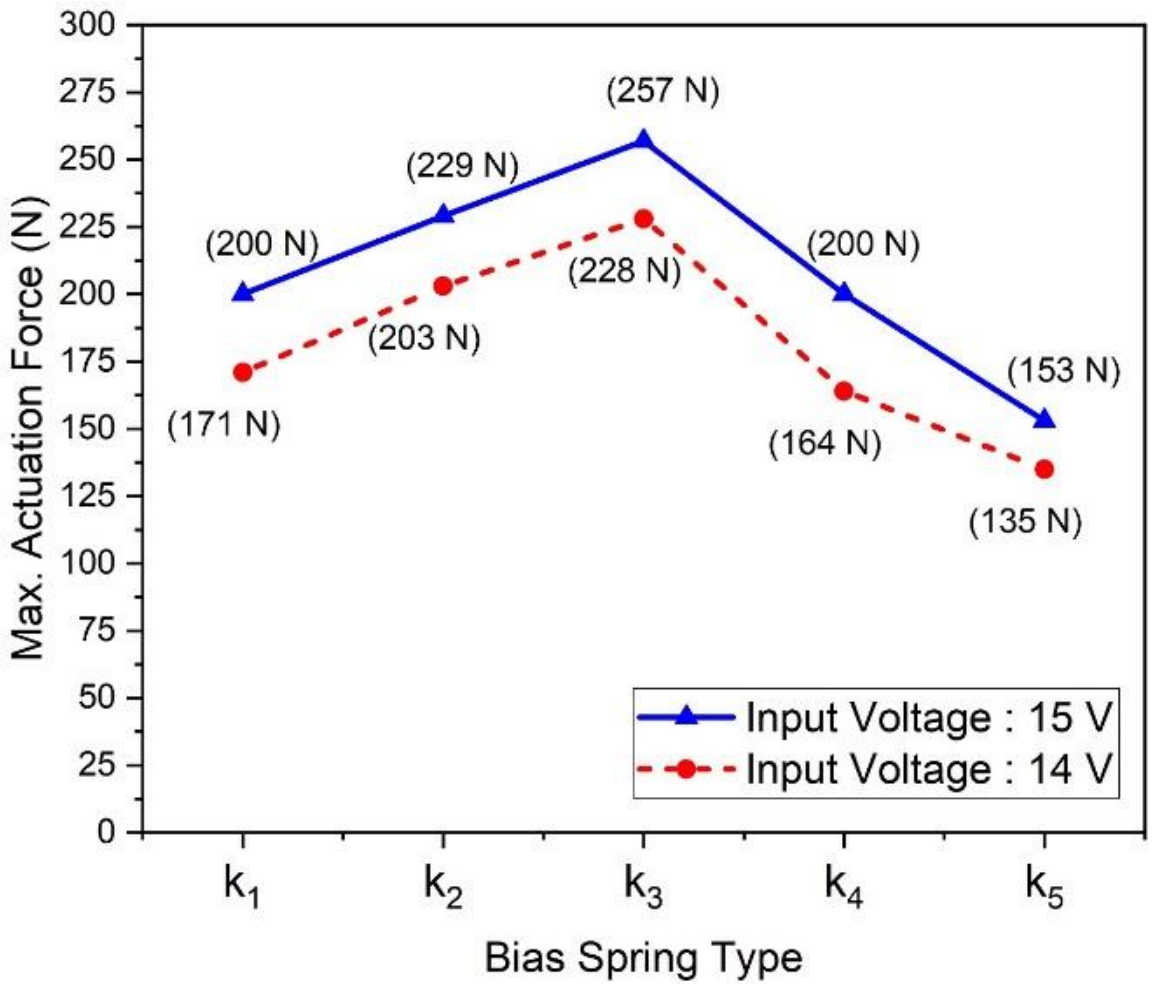

(b)

| **Bias spring type** | **Material type** (Standard) | **Active coils** (nos.) | **Inner diameter** (mm) | **Wire diameter** (mm) | **Stiffness** (N/mm) |
|---|---|---|---|---|---|
| $k_1$ | Stainless steel (BS 970) | 13.00 | 14.0 | 2.30 | 5.12 |
| $k_2$ | Stainless steel (BS 970) | 14.00 | 14.1 | 2.25 | 4.23 |
| $k_3$ | Steel wire (IS 4454-1) | 10.75 | 14.6 | 2.00 | 3.36 |
| $k_4$ | Steel wire (IS 4454-1) | 14.50 | 14.6 | 2.00 | 2.49 |
| $k_5$ | Steel wire (IS 4454-1) | 14.00 | 15.0 | 1.80 | 1.62 |

(c)

Figure 10 The plots demonstrate the relationship between actuation force, bias spring constant, and input voltage to determine the optimal design and operating conditions. (a) illustrates the plot of maximum actuation force (averaged based on 10 trials) and corresponding SMA wire temperature with respect to the input voltage. With the increase in voltage pulse, both the actuation force and SMA wire temperature increase. It was observed that 15 V for 10 seconds could be considered the optimal operating voltage to avoid overheating, in line with the data available from the DSC experiment. (b) highlights the dependence of the bias spring constant on actuation force for two different actuation voltages (14 V and 15 V). Initially, a negative correlation is observed between the spring constant and the force generated, indicating a decrease in the spring constant increases the force as less bias force is required to overcome. However, as the spring constant decreases further, the force decreases due to insufficient pre-tension in the SMA wires to generate maximum force. Additionally, experimental inference reveals that increasing the actuation voltage results in a higher actuation force. For the bias spring, k_{3}=3.36 N/mm is identified as the optimal bias spring constant. (c) presents the table with different compression spring types along with their design parameters, including material, number of active coils, inner diameter, wire diameter, and stiffness. (Reference for bias spring parameter data: Design and inspection sheet provided by the spring manufacturer, MESCO)

transformation of SMA. For the case of 10 seconds of actuation time, a 7% increase in SMA wire temperature was observed on increasing the voltage input from 13 V (98 °C) to 14 V (104 °C), and a 15% increase in SMA wire temperature was observed on increasing the voltage input from 14 V (104 °C) to 15 V (120 °C). It is imperative to note that this set of experiments was conducted with the bias compression spring as $k_1$ (5.12 N/mm). Thus, considering the force generated and the maximum SMA wire temperature, 15 V was considered as the operating input voltage.

Furthermore, the stiffness of the bias compression spring has been identified as one of the most influential design parameters on the output force of the actuator presented. The bias compression spring is essential to the actuation mechanism, as it provides the necessary pre-tension to the SMA wires and assists the actuator in returning to its initial position after actuation. It is crucial to note that a portion of the initial force generated by the actuator is used to overcome the resistance of the bias spring force. To ensure optimal actuation, it is essential to use an optimized bias spring that imparts sufficient tension to the SMA wires while restoring the actuator to its original position after actuation. The use of an appropriately designed compression spring would also facilitate the efficient use of the generated force, thereby optimizing the overall performance of the actuator system. The experimental data under two different input voltage conditions (14 V and 15 V) represented in Figure 10(b) demonstrates that a decrease in the spring constant increases the actuation force. However, the force output began to decrease after some time because the subsequent compression spring type with lower stiffness could no longer impart sufficient pre-tension in the wires for effective actuation. The experimental results illustrated in Figure 10(b) and Figure 10(c) indicate that a 34% decrease in the spring constant (from $k_1 = 5.10$ N/mm to $k_3 = 3.36$ N/mm) resulted in a 29% increase in the maximum force output, from 200 N to 257 N for 15 V input voltage case. However, when the spring constant was further reduced, beyond the optimal limit, by 51% (from $k_3 = 3.36$ N/mm to $k_5 = 1.62$ N/mm), the maximum force output began to diminish by approximately 40% (from 257 N to 153 N). Additionally, Figure 10(b) depicts a discernible increase in force output as actuation voltage increases. Notably, when the actuation voltage was maintained at 15 V for 10 seconds, the resulting force output was found to be 16% greater than when the actuation voltage was maintained at 14 V for the same actuation time. Thus, considering the force generated, $k_3$ was considered as the optimized bias compression spring.

## *4.3. Actuation Force*

### *4.3.1. Case study for input pulse of 15 V*

The model depicted in Figure 5 is experimentally investigated to measure the actuation force of the SMA-based bipennate linear actuator. The actuator model having bias spring constant ($k_3 = 3.36$ N/mm) without HSAM has been taken into consideration for this experimental study. As stated earlier, the shape memory effect (SME) occurs when a shape memory alloy material, in this case, a wire, reaches a phase transformation temperature threshold. To attain this elevated temperature, the current study employs the Joule heating method, which entails applying a voltage differential across the circuit in order to raise its temperature above the required threshold for actuation. As depicted in the Figure 5(e), one side of the central chassis frame is equipped with five horizontal (into the plane) SMA stacks and three vertical (on the plane) stacks of SMA wires, which are mirrored on the opposite side of the frame for symmetrical configuration. As shown in the Figure 5(a), the resistive circuit of the SMA wires architecture is such that the horizontal stacks are arranged in series while the vertical stacks are arranged in parallel electrical connections.

Figure 11 highlights a 15 V input voltage condition for 10 seconds, followed by 15 seconds of cooling interval, and this cycle was subsequently repeated twice. Throughout the two cycles, the actuator withdrew a peak current of 4.9 A, resulting in an maximum output power consumption of 73.5 W. In Figure 11(a), the temporal distribution of the temperature of the SMA wire is obtained by performing the simulation of the mathematical model in Mathworks Simulink R2020b, under the influence of an applied potential difference. During the first cycle, upon the application of the potential difference for ten seconds, the temperature of the wire rises and achieves a peak of 82°C. Following the removal of the potential difference, the temperature of the wire reduces and settles at 20°C, a value close to the ambient temperature of 18°C at the start of the first cycle. Correspondingly, in the subsequent heating cycle, the temperature of the wire rises again and reaches a maximum of 83°C. Figure 11(b) depicts the temporal profile of the blocked force generated by the actuator measured using a piezoelectric load cell (PACEline CFT/5kN). The force generated by the actuator is dependent on the heating and cooling cycles. Notably, the force exhibits a monotonically increasing trend during the heating phase, followed by a decreasing trend when the voltage supply is zero. Furthermore, after the first heating cycle ended, a maximum force output of 237 N was reported; following a 15-second cooling cycle, the force output decreased to 42 N. Subsequently, on the completion of the second heating cycle, a maximum force output of 257 N was obtained.

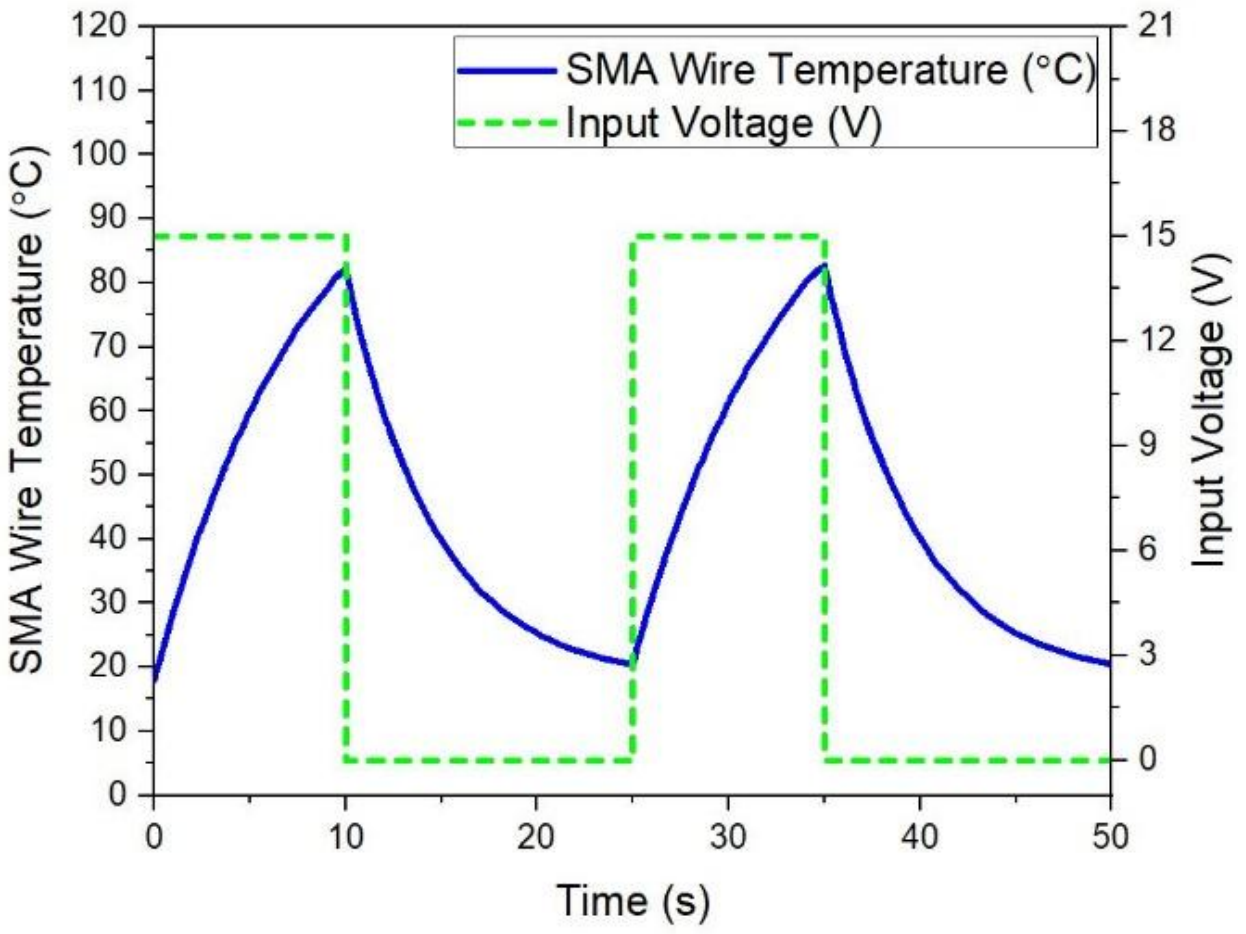

(a)

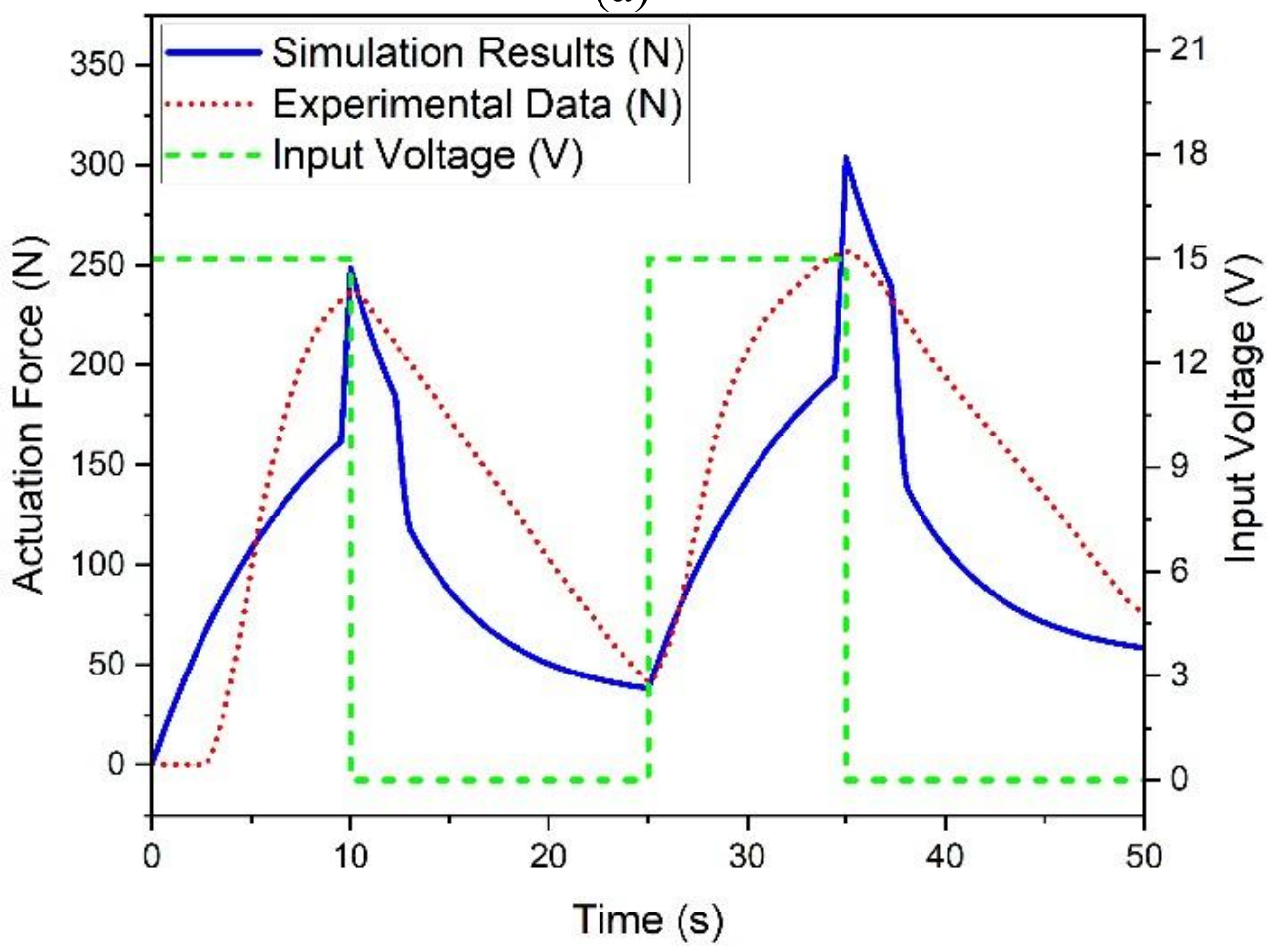

(b)

Figure 11 (a) illustrates the temporal distribution of the SMA wire temperature obtained by solving analytical model. The maximum SMA wire temperature obtained after the first and second cycle was 82°C and 83°C, respectively. (b) depicts the temporal distribution of the force generated by the bipennate SMA-based linear actuator without HSAM. The actuation process involved two cycles of heating and cooling, achieved through the Joule heating phenomenon. In each cycle, the actuator received an input of 15 V during the heating phase, lasting 10 seconds. Subsequently, it was allowed to cool for the next 15 seconds. The experimental maximum force at the end of the first cycle was measured as 237 N, while the force reached 257 N at the end of the second cycle. The graph also provides a comparison study of the actuation force obtained from experimentation and analytical model.

A comparative study between the actuation force data obtained from experimenation and analytical model shows that the maximum force obtained via simulation (249 N) was 5% more than the observed experimental force (237 N) for the first cycle. Similarly, for the second cycle, the maximum force obtained via simulation (304 N) was 18% higher than the observed experimental force (257 N). The frictional losses in

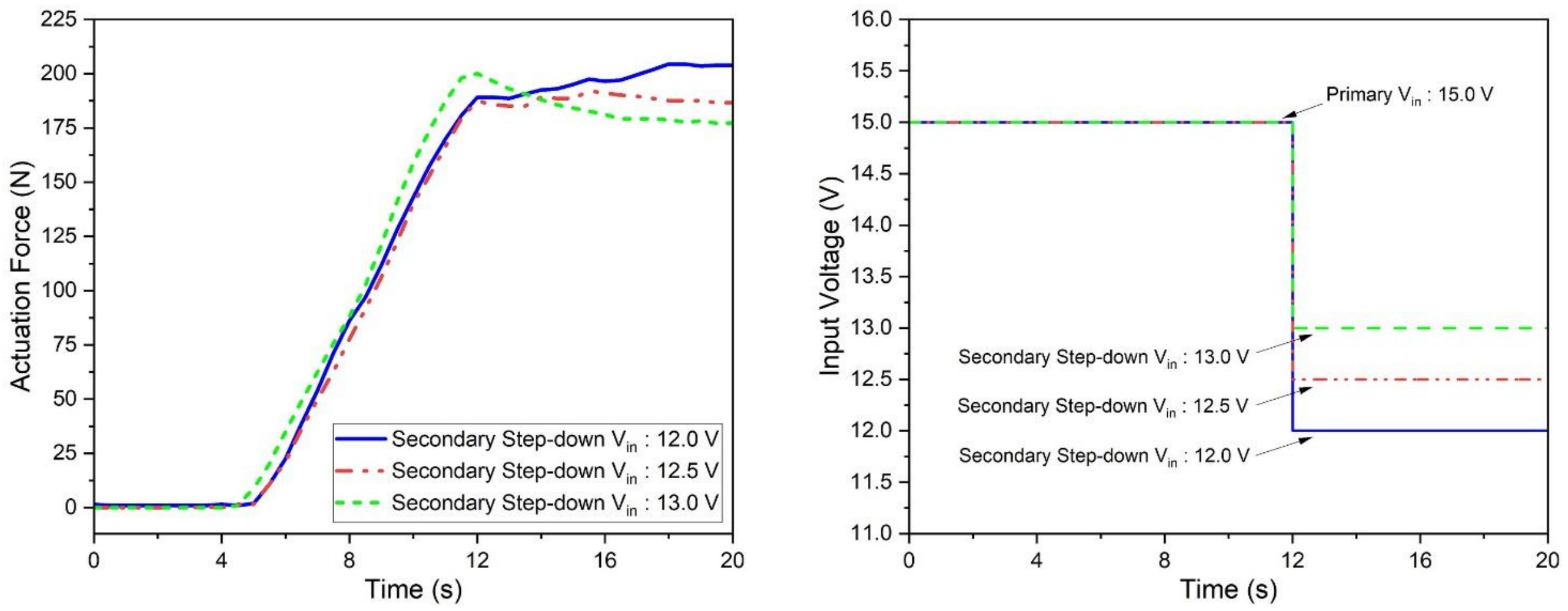

(a) (b)

| Test cases | Actuation force at $t = 12$ sec ($F_{act\|t_{12}}$) | Mean sustained actuation force ($F_{m\|sustained}$) | Standard deviation of sustained actuation force | $F_{m\|sustained} - F_{act\|t_{12}}$ |
|---|---|---|---|---|
| Secondary input voltage: 12.0 V | 189.1 N | 197.1 N | 5.6 N | 7.9 N |
| Secondary input voltage: 12.5 V | 187.6 N | 188.1 N | 1.9 N | 0.5 N |
| Secondary input voltage: 13.0 V | 199.9 N | 183.6 N | 6.6 N | -16.4 N |

(c)

Figure 12 The plot demonstrates the generation of sustained actuation force under the two-step driving voltage strategy. (a) depicts the actuation force generated by the actuator is shown for three different secondary step-down input voltages. Initially, a maximum force is achieved by applying the primary voltage, which is then sustained by stepping down to the secondary voltage. (b) displays the input voltage during the three different iterations performed to maintain the sustained actuation force. The total actuation cycle lasted for 20 seconds. During the first 12 seconds, the actuator was subjected to a primary voltage of 15 V. For the remaining 8 seconds; three secondary step-down voltages were applied as the input voltage (12.0 V, 12.5 V, and 13.0 V). (c) The table illustrates different statistical parameters of the sustained actuation force experiment. The force was reported at t=12 seconds to highlight the maximum actuation force generated at the end of applying the primary input pulse. The mean and standard deviation of the sustained actuation force depict the respective parameters of the force generated solely during the secondary step-down voltage cycle. The average difference between force at t=12 seconds and sustained actuation force measures how close the sustained actuation force in the secondary step-down voltage cycle is to the maximum force generated due to primary voltage.

the system during the actuation are responsible for this difference. The analytical model developed was observed to align with the experimental finding and data fell within an acceptable tolerance range. These results demonstrate the potential utility of the linear actuator in applications requiring high-force actuation output.

### *4.3.2. Case study for variable input voltage condition*

In the earlier section, it was observed that the temperature of the wire increased considerably via the Joule heating phenomenon. Due to the limited actuation time, a constant voltage pulse for 10 seconds resulted in a sharp rise in SMA wire temperature, indicating rapid heat generation. Maintaining the SMA wires within a safe operating temperature limit is crucial to prevent overheating. Exceeding this limit can lead to dislocation movement within the lattice structure of the wires, resulting in the degradation of memory recovery capabilities. Nonetheless, many practical applications necessitate the application of sustained actuation force over an extended period of time, mandating the development of a suitable strategy to meet this demand. One approach to address this challenge is by implementing a two-step driving current strategy. This method involves initially applying a larger current (or voltage) to heat the SMA wire, followed by reducing the current (or voltage) to maintain its temperature without overheating. This allows for precise control of the temperature of the SMA wire and helps prevent any detrimental effects caused by excessive heating.

Figure 12(b) depicts the two-step driving voltage strategy, with a primary input voltage of 15 V for 12 seconds and three secondary step-down inputs of 12 V, 12.5 V, and 13 V for three separate test cases. The primary voltage input aided in establishing the initial conditions necessary for the subsequent application of sustained actuation force and, thus, meeting the actuation force requirement. Subsequently, a secondary voltage was applied for 8 seconds, allowing the actuator to maintain the necessary force for the duration specified. This method is particularly advantageous for applications in which a sustained actuation force must be maintained for a predetermined amount of time, such as operating mechanized limbs or activating mechanical switches. Using a combination of primary and secondary voltage inputs, the actuator can sustain the requisite actuation force output for an extended period of time, making it suitable for a wide range of practical applications.

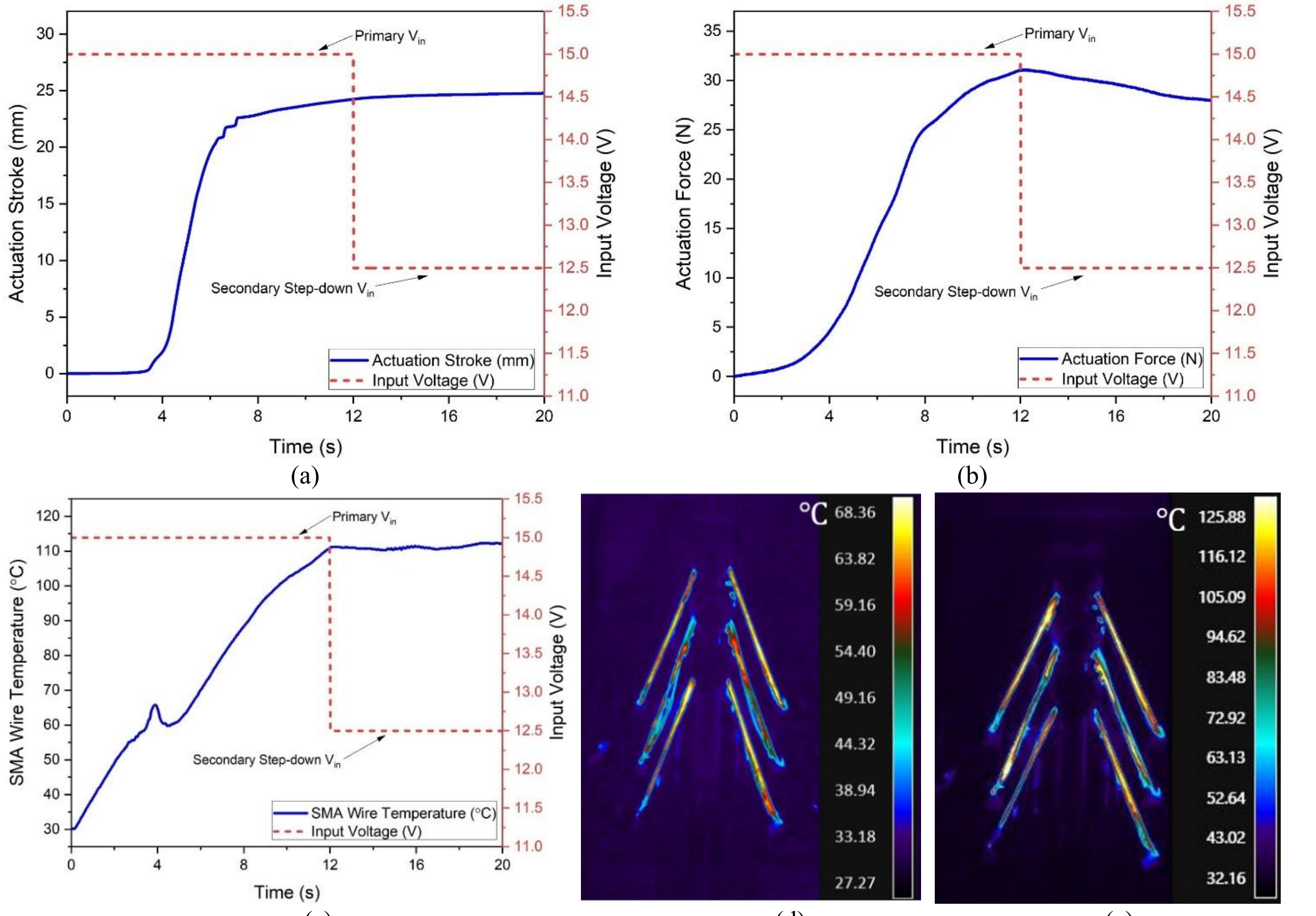

Figure 13 The plots represent actuation force and stroke measurement experiments conducted on a medium force-medium stroke actuator (with HSAM) under thermal monitoring. Initially, a 15 V primary input voltage was applied for the first 12 seconds, followed by a secondary step-down voltage of 12.5 V for the remaining 8 seconds. (a) highlights the temporal distribution of the stroke of the actuator. (b) illustrates the force generated by the medium force-medium stroke actuator. (c) depicts the temperature profile variation with respect to time. Under the action of primary voltage, the SMA wire temperature increases; however, it stabilizes when the secondary step-down voltage is applied. Additionally, (d) and (e) present thermal images captured at t = 4 sec and t = 12 sec instances, respectively, using the FLIR Research Studio software and FLIR A700 thermal imaging camera. These images highlight the temperature distribution in the SMA wires of the bipennate-based linear actuator. (Refer to the supplementary section for working demonstration video of the actuator integrated with the stroke amplifier).

The result of the actuation force subjected to a two-step driving voltage condition is plotted in Figure 12(a). Under the action of primary voltage (15 V), the average force measured for the three different experiments (with different secondary step-down voltages) after the application of primary voltage (at 12 secs) was 192.2 N, with a standard deviation of 6.8 N. The application of secondary voltage was intended to achieve a more consistent and stable force. When a secondary voltage of 12 V was applied, a sustained actuation force of 197.1 N was observed, with an average difference of 7.9 N between the sustained actuation force and the maximum force with a standard deviation of 5.6 N. Similarly, in the case of a 12.5 V step-down secondary input pulse, an average sustained actuation force of 188.1 N was observed, with an average difference of 0.5 N between the sustained actuation force and the maximum force, along with a standard deviation of 1.9 N. In the last case, 13 V was chosen as the secondary step-down pulse, and an average sustained actuation force of 183.6 N was observed, with an average difference of -16.4 N between the sustained actuation force and the maximum force having a standard deviation of 6.6 N. Based on the statistical evaluation of all three cases and calculating the mean difference and standard deviation of the measured data as tabulated in Figure 12(c), a combination of a primary voltage of 15 V and a secondary step-down pulse of 12.5 V was considered to be the optimal two-step driving voltage condition.

| Parameter | 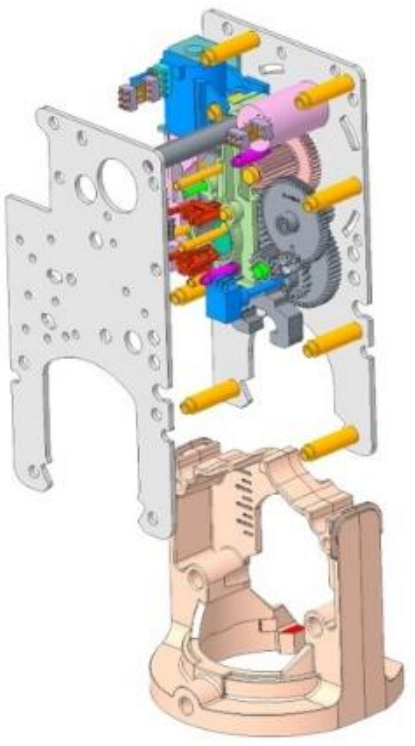 **VA-4233 Linear Actuator** (Stepper motor driven) | 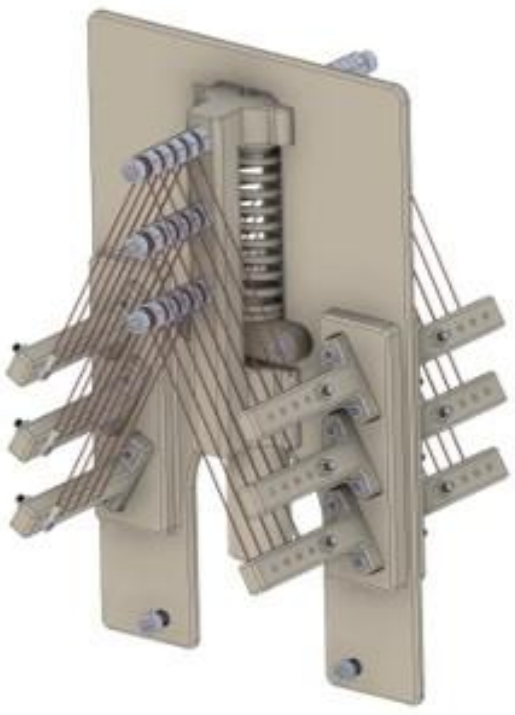 **Bipennate Muscle - Linear Actuator** (Shape memory alloy driven) | **% Change** |
|---|---|---|---|
| Weight of drive mechanism (Only) | 0.9 kg | 0.3 kg | 67 % ↓ |
| Envelope dimension | 173 mm × 104 mm × 91 mm | 165 mm × 100 mm × 101 mm | 1.7 % |
| Number of components | 60 | 12 | 80% ↓ |
| Cost of drive mechanism (Only) | \$40 | \$27 | 32% ↓ |
| Actuation energy consumption | 912 J | 840 J | 7.9 % ↓ |

Table 5 A comprehensive comparative analysis of a bipennate SMA-based actuator vis-à-vis a DC stepper motor gear-train-based conventional actuator (VA-4233 linear actuator). The analysis focuses on performance benchmarking of various parameters, taking into consideration their respective driving mechanisms.

### *4.4. Actuator integrated with stroke amplification mechanism*

The modular actuator developed can be configured as a high force-low stroke or a medium force-medium stroke actuator, depending on the stroke amplification mechanism required. A hydraulic stroke amplification mechanism (HSAM) with a stroke amplification factor of 11.8 is integrated, which can amplify the stroke length by 11.8 times and also reduces the actuation force output by the same factor. The experiments were conducted to evaluate the performance of the actuator with and without the HSAM, assessing parameters such as actuation force and stroke output, along with thermal data monitoring. The experiment conducted without the HSAM resulted in a significantly higher force output than with the HSAM.

As depicted previously in Figure 3(f), the integration of a hydraulic stroke amplifier substantially increases the actuation stroke. Figure 13 illustrates that the maximum actuation stroke has increased to 24.8 mm, while the maximum actuation force has decreased to 31.1 N for the two-step driving voltage of 15 V (primary) and 12.5 V (secondary). With the incorporation of HSAM, the linear actuation design can be classified as a medium-force, medium-stroke actuator, resulting in a reduction in force with an increase in stroke compared to a high force-low stroke actuator design. Nonetheless, this conversion is necessary for applications requiring medium force and stroke. The input voltage applied to the actuator was 15 V for 12 seconds and 12.5 V for the remaining 8 seconds. As evident from Figure 13(b), during the first 12 seconds, the actuator output steadily increased to 31.1 N, indicating the quick responsiveness of the actuator to the input voltage. It can be further observed that the actuator force value nearly sustained close to the maximum upon applying secondary voltage input. The average force generated by the actuator under the application of the secondary step-down voltage of 12.5 V was 29.5 N. The medium stroke generated by the system is suitable for applications requiring an extended stroke, making it a desirable characteristic of the actuator. In conclusion, incorporating a hydraulic stroke amplifier significantly increases the actuator stroke while reducing the actuator force to a medium level, making it suited for various applications.

The reliability of the actuator is demonstrated by its responsiveness to the input voltage and its ability to sustain actuation force close to the maximum force upon the application of secondary voltage input. Throughout the experimentation, rigorous thermal monitoring was carried out to avoid overheating and maintain the integrity of the SMA wires. During the initial 12 seconds following the application of the primary voltage, the wire temperature exhibited a continuous and monotonic increase while supplying a constant input voltage. Figure 13(c) shows the SMA wire temperature variation with respect to time, wherein it can be observed that the SMA wire temperature reached 111.13 °C at the end of the primary voltage phase. Upon the subsequent application of the secondary step-down pulse, the SMA wire temperature attained stability. After the initial 12 seconds, the temperature exhibited a mean of 111.13 °C and a standard deviation of 0.53 °C. Figure 13(d) and Figure 13(e) showcase the thermal image of the SMA-based bipennate linear actuator captured using FLIR A700 thermal imaging camera at $t = 4$ sec and $t = 12$ sec instances, respectively. These results demonstrate the consistency and reliability of the thermal behavior of the SMA wire during the experimental investigations and confirm the safe operation of actuation without overheating.

### *4.5. Performance benchmarking with VA-4233 actuator*

VA-4233 is a linear actuation device from Johnson Controls that utilizes a stepper motor to position control valves precisely in HVAC applications. The electric valve actuators of the VA-4233 Series provide a high force output for floating, on/off, or proportional control. It can be field-mounted or factory-coupled without additional linkage to Johnson Controls VG7000 Series bronze control globe valves. In addition, these actuators can be field-mounted to select valves using Johnson Controls' mounting packages. A comparison has been made between the currently developed bipennate-based SMA linear actuator and an actuator with similar fitting features.

As illustrated in Table 5, it has been observed that the VA-4233 actuator generated an actuation force of 271 N. In comparison, the novel SMA actuator described in current research generated a force of 257 N, which is within the acceptable tolerance band of the actuation operation. The bio-inspired shape memory alloy (SMA) actuator system exhibited a 67% reduction in the weight of the driving mechanism and weighed approximately 0.3 kg as compared to the 0.9 kg of the VA-4233 model. The SMA actuator showcased a simplified modular design compared to the VA-4233 actuator. While the VA-4233 required 60 components, the SMA actuator was able to achieve the same functionality with just 12 components. This reduction in the number of required components signifies an 80% decrease in complexity. Additionally, fewer components contribute to increased reliability, as there are fewer potential points of failure or maintenance requirements. The reduced complexity also aids in enhancing the overall durability of the actuator. Furthermore, the cost of the drive mechanism of VA-4233 and the SMA actuator estimated at the mass production level highlighted the cost saving by 32% in the latter case. A comparasion of the total energy requirement for the two cases has been made. While SMA requires high power (84 W) for a smaller time duration (10 sec), the stepper motor-based system requires lower power (12 W) for a longer duration (76 sec), resulting in about 8% energy savings in the case of SMA-based actuation system. However, this may be further improved with a more controlled energy dissipation which will be addressed in our future work. The novel bipenniform configuration-based SMA actuator developed in the present research demonstrated a comparable actuation force output to the existing VA-4233 actuator while offering significant advantages in terms of weight reduction, lesser components, cost and energy savings, and similar envelope dimensions for assembly compatibility with dampers and louvers.

## Introducing a bioinspired curved fibre SMA actuator for enhanced stroke

The earlier sections investigated the development and performance of two bioinspired SMA actuator configurations. These comprised a high-force, low-stroke actuator without HSAM and a medium-force, medium-stroke actuator incorporating HSAM. The straight bipennate arrangements considered in the preceding sections correspond to a single sloped bipenniform configuration, in which the SMA wire follows a single inclination relative to the actuation direction. The configuration introduced in the present section extends this concept to a multisloped bipenniform arrangement, in which successive changes in the inclination of the SMA wire are introduced along its active length. The present section explores a variation in the geometric arrangement of the SMA wires intended to increase the attainable stroke while retaining a high force output comparable to that of high force – low stroke (without HSAM). In the proposed configuration, the SMA wire is reeved over a set of pins to create an extended and curved fibre configuration. Unlike the earlier configurations, in which the SMA wires provided an engineering approximation of the muscle fibres in a bipennate architecture, the reeved arrangement enables a fibre trajectory that more closely represents the curved arrangement observed in biological muscle structures. More importantly, the extended wire path allows a greater active SMA wire length to be accommodated within the actuator geometry. In a straight bipennate arrangement, the required stroke is largely constrained by the available straight SMA wire length and therefore constrains the overall dimensions of the actuator. By routing the wire over pins, the width of the actuator arrangement can be reduced while retaining an active wire length capable of producing a stroke comparable to that of a straight bipennate arrangement occupying a larger envelope dimension. It therefore offers a geometric design platform for tailoring the actuator dimensions according to the required linear stroke and for achieving increased actuation within a compact structure while retaining comparable force capability.

This section examines the initial development of the reeved SMA wire arrangement intended to provide a relatively higher stroke at comparable force levels. Figure 14(a) shows the geometric arrangement of the SMA wire in the straight bipennate configuration in its initial and actuated states. A single SMA wire was used in the present configuration to demonstrate and validate the underlying geometric concept as an initial proof of concept. Extending this behaviour to multiple SMA elements follows directly from the already established stacked bipennate configuration presented earlier in Figure 3, where additional SMA elements are incorporated to achieve higher force output, and therefore demonstrating the present single-wire response inherently supports the feasibility of the stacked implementation as well. Figure 14(b) shows one possible form of the proposed configuration, in which the SMA wire is reeved over two pairs of pins. In the corresponding single sloped arrangement, reducing the lateral dimension by directly moving the SMA wire attachment point closer would also reduce the bipennate angle and alter the force transmission geometry. In the proposed multisloped arrangement, the first pin is therefore positioned such that the initial wire segment retains approximately the same bipennate angle as the corresponding single sloped configuration. This preserves a comparable force transmission geometry, while the subsequent routing of the wire over additional pins accommodates a greater active SMA wire length within a reduced lateral envelope. Geometric analysis indicated that this arrangement increases the effective active wire length accommodated between the pins $P_1$, $P_2$ and the bias spring attachment point compared with the corresponding straight bipennate arrangement. The greater active wire length results in a larger accumulated contraction during SMA transformation, which contributes to the net displacement of the bias spring attachment. Consequently, the multisloped arrangement enables a reduction in lateral space while retaining a displacement of the same order as the approximately 2.1 mm response obtained from the corresponding single sloped single wire arrangement discussed in Section 4.4. During SMA transformation, contraction along this extended wire path contributes to the displacement of the bias spring attachment, thereby increasing the available actuator stroke.

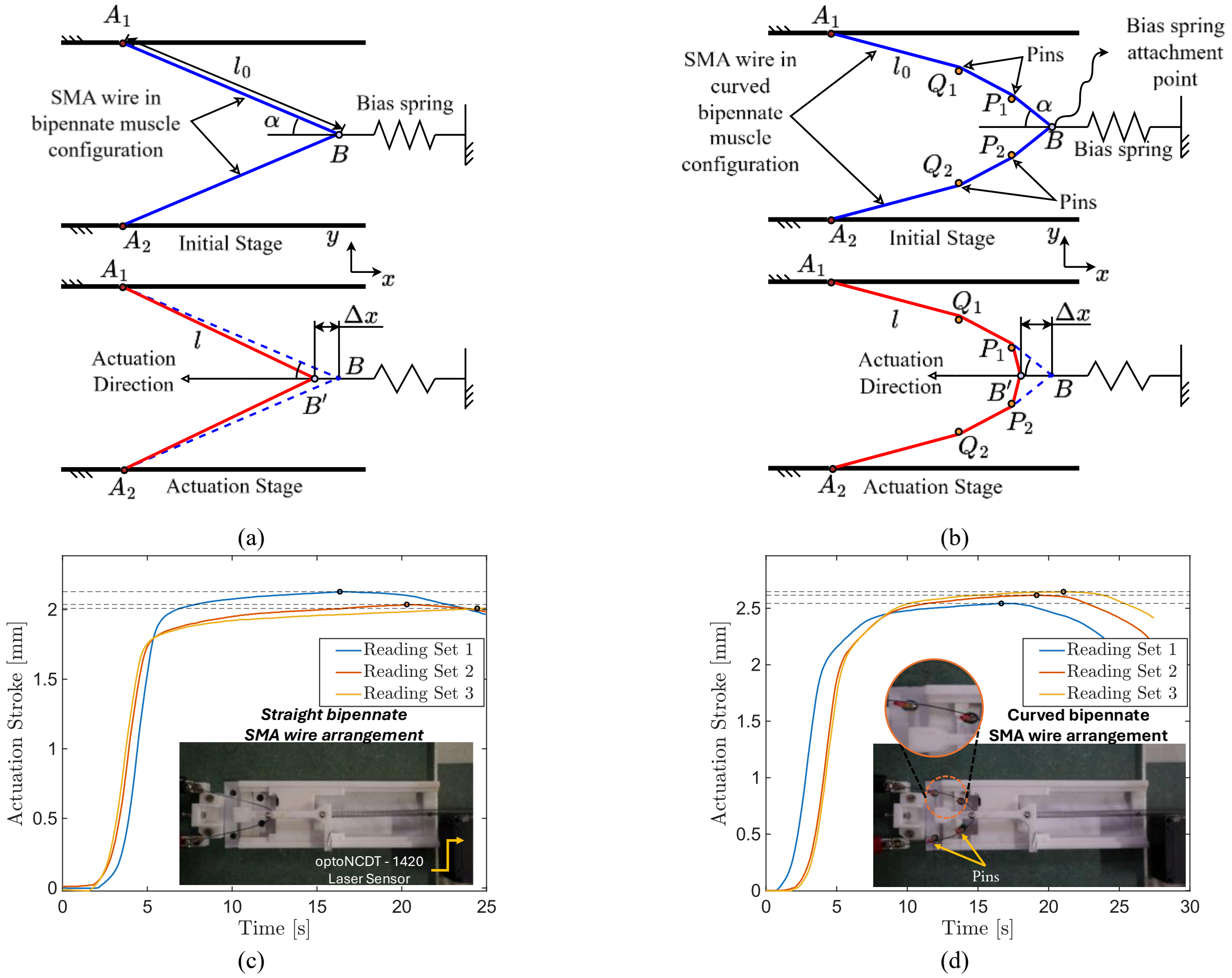

Figure 14 Demonstrates the geometric configuration and stroke response of the straight and curved bipennate SMA wire arrangements. (a) shows the single sloped bipenniform architecture (SSBA) in the initial and actuated states, represented by blue and red lines, respectively. (b) presents the corresponding multisloped bipenniform architecture (MSBA), where routing over the pins allows contraction along the ($A_1$) to ($P_1$) segment to contribute to the displacement transmitted between ($P_1$) and ($B$). (c) presents the experimental setup and stroke response of the SSBA over three independent trials, with a mean maximum stroke of 2.058 mm. (d) presents the corresponding results for the MSBA, which achieves a mean maximum stroke of 2.602 mm. The experimental comparison demonstrates the increased stroke capability of the curved SMA wire arrangement and supports the corresponding geometrical analysis.

The prototype was supplied with a current of 2 A for ~20 s to activate the SMA wire. The actuation stroke was measured using an optoNCDT-1420 laser sensor. The corresponding experimental responses are presented in Figure 14(c) and (d). Based on three independent experimental trials,

| Parameter | Single sloped configuration | Multisloped configuration | Comparison |
|---|---|---|---|
| Wire geometry | Straight bipennate path | Curved multisloped path | Modified |
| Initial bipennate angle | Reference angle | Approximately maintained | Same |
| Active SMA wire length | Lower | Higher | Increased |
| Lateral space requirement | Higher | Lower | Reduced |
| Mean maximum stroke | 2.058 ± 0.063 mm | 2.602 ± 0.053 mm | Increased |

Table 6 A comprehensive comparative analysis of the single sloped bipenniform architecture (SSBA) and the multisloped bipenniform architecture (MSBA). The analysis focuses on the influence of geometric arrangement on actuator characteristics, including active SMA wire length, lateral space requirement and actuation stroke

the proposed curved bipennate arrangement achieved a mean maximum stroke of 2.602 ± 0.053 mm, whereas the straight bipennate arrangement produced 2.058 ± 0.063 mm, where the values represent mean ± standard deviation. This corresponds to an approximately 26.4% increase in the mean maximum stroke and is consistent with the trend predicted by the geometrical analysis. These results demonstrate the feasibility of using the geometric routing of the SMA wire as an additional design parameter for increasing the attainable actuator stroke. The configuration consequently provides a potential route towards compact SMA actuators in which the required linear displacement can be achieved without proportionally increasing the lateral dimensions of the actuator.

## Introducing SMA coil-based high force-high stroke design actuator

In the earlier sections, the present research work explored the development and performance assessment of two distinct types of bio-inspired SMA actuators: high force - low stroke (without HSAM) and medium force - medium stroke (with HSAM). The choice of the actuator design depends on the intended application, and hence it could serve a variety of purposes as per the requirement. In this section, we introduce an alternative design strategy utilizing shape memory alloy (SMA) coils that could be used as a high force - high stroke actuation design. Utilizing the unique characteristics of SMA coils, such as the shape memory effect and pitch angle geometry, the system design generates a substantial displacement without incorporating any external stroke amplification mechanism. The geometry of the SMA coils is defined with SMA wires arranged in a helical pattern, allowing it to generate the maximum displacement along the axis of the coil from the given strain rate. Such configuration provides a design platform to optimize the actuation system with respect to the linear stroke requirement. Therefore, actuators incorporating SMA coils can accomplish a higher stroke than conventional SMA wires without compromising the actuation force output.

This section focuses on developing an actuator prototype with a high force - high stroke design configuration. Figure 15(a) and Figure 15(b) show the top and front view of the SMA coil-powered bipenniform configured linear actuator. The electrical circuit comprises two sets (A & B) of four springs connected in series, while the four springs within each set are connected in parallel, as shown in Figure 15(c). This configuration assures a uniform current distribution to the individual SMA element. The prototype has been supplied with a 12 A current for 15 seconds to provide heating of the SMA coils. The resistive circuit design ensures that each SMA coil element is thus supplied with 3 A of current. Additionally, it has been observed that the pitch angle, which is the angle between adjacent coils, has a strong inverse correlation with the actuator stroke. The transformation strain of the SMA material leads to a reduction in the pitch angle resulting in a considerable increase in the stroke of the SMA actuator. The measurement of actuation force and stroke of the system are being recorded using PACEline CFT/5kN Load Cell and optoNCDT 1420 Laser Sensor, respectively. The experimental findings are presented in the Figure 15(d) and Figure 15(e) demonstrate the efficacy of the high force - high stroke actuator. The actuator exerts a force that steadily increases to a maximum of 102 N as illustrated in Figure 15(d). The system achieves a maximum stroke of 64 mm and exhibits the temporal distribution, which follows a monotonically ascending trend until the SMA coil attends its solid length, as shown in Figure 15(e).

## Conclusion

Pennate muscles are well-known to provide the  advantage of generating high force in a quasi-static mode in a confined physiological region. This paper presents a novel bio-inspired shape memory alloy-powered linear actuator wherein the design configuration follows bipennate muscle architecture.

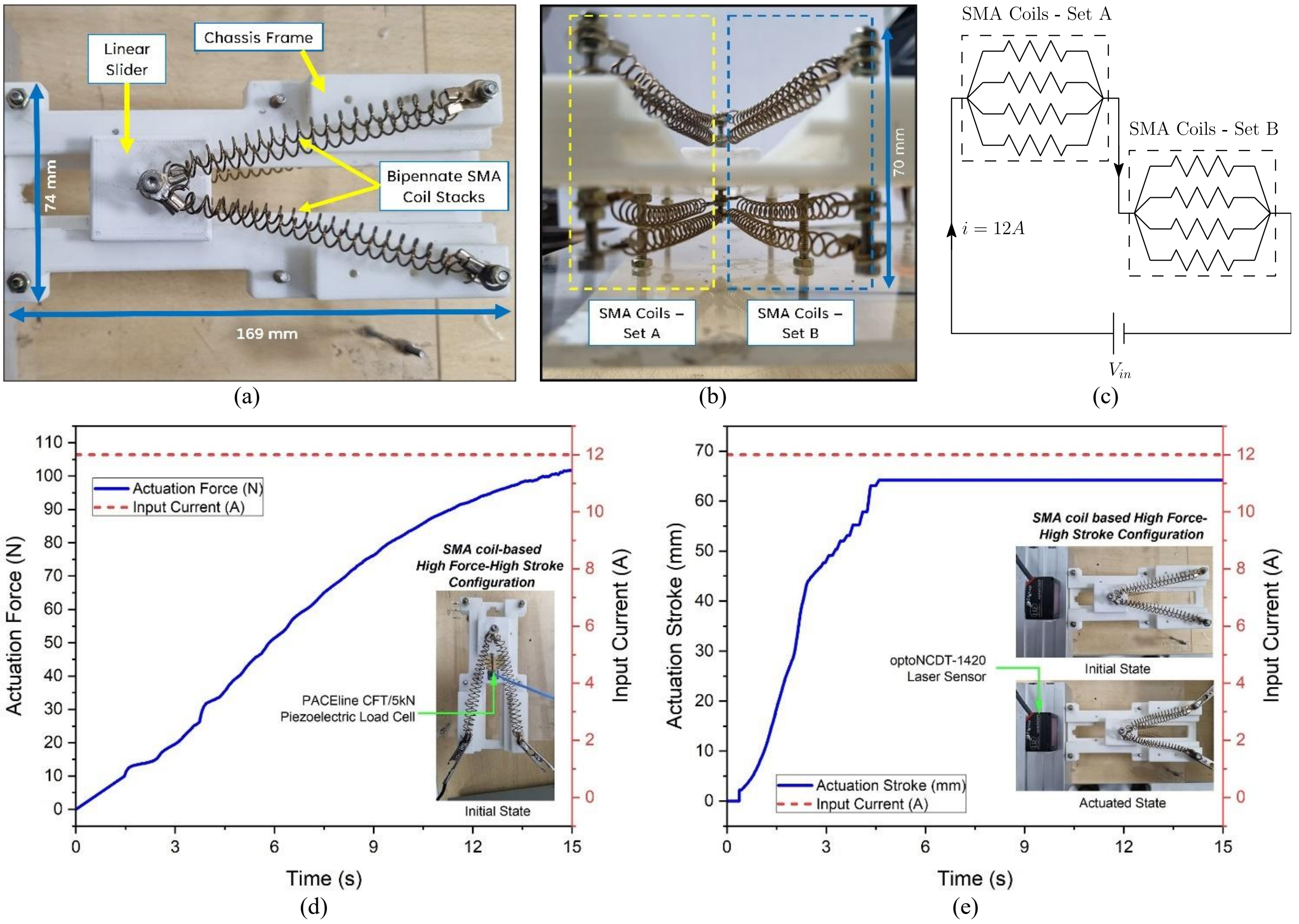

Figure 15 Demonstrates design and performance measurement of the SMA coil-based linear actuator as a high force - high stroke actuation system. (a) shows the top view of the system, highlighting the chassis frame, linear slider, and bipennate SMA coil stacks. (b) presents the front view of the system depicting the two sets of SMA coil stacks connected in the parallel and series combination. (c) shows the electrical resistive circuit where sets A and B are connected in series. (d) presents the temporal distribution of the force generated by the coil-based actuator. The plot illustrates that the force monotonically increases throughout the actuation process, reaching a maximum of 102 N after 15 seconds. (**e**) showcases the stroke variation with respect to time. It depicts the pattern of the stroke of the actuator, which continuously increases until the SMA coil reaches its solid length generating a stroke output of 64 mm. The experimental data highlights the capability of the actuator design to achieve high displacement or stroke

The compatibility of the modular actuator with the hydraulic stroke amplification mechanism has been explored. The paper discusses and investigates **four** distinct types of actuators design categories, a high-force-low-stroke system (SMA wires without any amplification mechanism), a medium-force-medium-stroke system (SMA wires with an amplification unit), an enhanced stroke system employing a multisloped bipenniform SMA wire arrangement (without an amplification unit) and a high-force-high-stroke system (SMA coils without any stroke amplification). A mathematical model was developed for multi-layered bipenniform configuration-based SMA actuator to solve the set of implicit governing equations using MathWorks Simulink environment. Designed using only 12 components as compared to 60 components in a similar traditional actuation system, the drive mechanism of this system offers the reliability of the actuation operation while reducing the production and maintenance cost of the unit significantly. The simplified design of the SMA actuator enhances its efficiency and makes it easier to optimize and fine-tune performance parameters. A design failure mode and effects analysis (DFMEA) was performed to identify potential failure modes and implement mitigation strategies. The model was fabricated using additive manufacturing techniques, and experiments were conducted to measure the actuation force, stroke, and thermal data of the system. Experimental data of the output force were consistent with the simlation results thus validating the mathematical model of the system. Performance assessments of the novel SMA actuation system were also benchmarked against the industry-developed stepper motor-driven actuator (VA-4233). The system has shown promising results generating an actuation force of 257 N with 15 V input voltage, meeting the acceptable range for actuation operation.

The current research work is partially motivated by the demands from the HVAC building automation and controls industry to develop smart material-based actuation systems. The present study explored the novel SMA driving principle and developed an alternative and effective solution for electromagnetism-driven conventional linear actuators coupled with multi-stage geartrain mechanisms. The developed system offers significant advantages in terms of 67% weight reduction, 80% lesser components, 32% cost reduction, 8% energy savings, and similar envelope dimensions for assembly compatibility with dampers and louvers for easy onsite installation. The utilization of additive manufacturing techniques for the fabrication of the actuator enables the rapid deployment of the system. The bipennate muscle-based SMA linear actuator has a wide range of applications, from building automation controls to lightweight actuation systems for space robotics and medical prosthesis. As a part of future development, the work will be focused on optimizing the multisloped bipenniform architecture and developing a high force-high stroke size-optimized compact system with a standalone controller and undertaking experiments for actuator lifecycle assessment. The research will also focus on implementing innovative evaporative cooling techniques on SMA elements using liquid coolants, which will further provide scope for increasing the actuation frequency and energy efficiency of the SMA-driven system.

## Acknowledgements

The authors would like to thank Abhishek Kumar Singh, Gyanendra Prakash Tripathi, and Anil Kumar Jha, who provided experimental insight and manufacturing expertise that greatly assisted the research.

## Declaration of Conflicting Interests

The author(s) declared no potential conflicts of interest concerning the research, authorship, and/or publication of this article.

## Funding

This research work has been supported by Johnson Controls India Pvt. Ltd. through a Corporate Social Responsibility (CSR) grant (Project Number: JCIPL/DORA/2021044).